\documentclass{article}

\usepackage{microtype}
\usepackage{graphicx}
\usepackage{booktabs} 

\usepackage{hyperref}

\usepackage[accepted]{icml2025}

\usepackage{amsmath}
\usepackage{amssymb}
\usepackage{mathtools}
\usepackage{amsthm}

\usepackage[capitalize,noabbrev]{cleveref}

\theoremstyle{plain}

\theoremstyle{definition}

\theoremstyle{remark}

\usepackage[textsize=tiny]{todonotes}

\usepackage{subcaption}
\usepackage{placeins}

\icmltitlerunning{Performance Plateaus in Inference-Time Scaling for Text-to-Image Diffusion Without External Models}

\begin{document}

\onecolumn
\icmltitle{Performance Plateaus in Inference-Time Scaling\\for Text-to-Image Diffusion Without External Models}




\begin{icmlauthorlist}
\icmlauthor{Changhyun Choi}{IPAI}
\icmlauthor{Sungha Kim}{IPAI}
\icmlauthor{H. Jin Kim}{IPAI,aero,others}
\end{icmlauthorlist}

\icmlaffiliation{IPAI}{Interdisciplinary Program in Artificial Intelligence, Seoul National University}
\icmlaffiliation{aero}{Aerospace Engineering, Seoul National University}
\icmlaffiliation{others}{ASRI, AIIS, Seoul National University}

\icmlcorrespondingauthor{H. Jin Kim}{hjinkim@snu.ac.kr}

\icmlkeywords{Machine Learning, ICML}

\vskip 0.3in


\printAffiliationsAndNotice{}  

\begin{abstract}
Recently, it has been shown that investing computing resources in searching for good initial noise for a text-to-image diffusion model helps improve performance. However, previous studies required external models to evaluate the resulting images, which is impossible on GPUs with small VRAM. For these reasons, we apply Best-of-N inference-time scaling to algorithms that optimize the initial noise of a diffusion model without external models across multiple datasets and backbones. We demonstrate that inference-time scaling for text-to-image diffusion models in this setting quickly reaches a performance plateau, and a relatively small number of optimization steps suffices to achieve the maximum achievable performance with each algorithm.
\end{abstract}

\section{Introduction}
\label{introduction}

In large language models, an inference-time scaling method has attracted a lot of attention because it has been discovered that this can improve performance without increasing the model size \cite{o1, guo2025deepseek, zhang2025and}. Therefore, studies have recently been conducted on whether inference-time scaling can also be applied to diffusion models \cite{ma2025inference, li2025reflect, zhuo2025reflection}. An influential study \cite{ma2025inference} in this area demonstrated that allocating more compute to search for a better initial noise helps to improve performance on text-to-image (T2I) tasks. Despite these advances, these studies all rely on additional vision-language models (VLMs) or other external models to evaluate the quality of the generated images. This makes it difficult to implement on consumer-grade GPUs typically found in personal desktops rather than in well-funded research laboratories or enterprises.

Independently, several approaches \cite{meral2024conform, guo2024initno, qiu2024self} have been proposed to optimize the initial noise and intermediate outputs of the denoising process at inference time so that the generated image aligns better with the input prompt. These methods do not require additional models; instead, they optimize the noise using only a pre-trained T2I diffusion model. In particular, the authors of InitNO \cite{guo2024initno} argue that not every noise sample drawn from a standard normal distribution precisely adheres to a given text prompt, implying the existence of both valid and invalid noise. This raises an important question: 
\begin{center}
\emph{When selecting good initial noise solely with the T2I diffusion model,\\does applying inference-time scaling help improve performance?}
\end{center}
We perform extensive experiments across various combinations of algorithms and models at multiple scales and reveal a clear performance plateau as we invest more computational resources, which contradicts both the default hyperparameter settings and common expectations. Based on these findings, we show how much computational effort should ideally be invested without relying on additional models when performing T2I tasks on GPUs with small VRAM. Furthermore, we show that the state-of-the-art (SOTA) algorithm among these initial noise optimization algorithms changes depending on the underlying T2I diffusion model, thereby indicating opportunities for future research.

\section{Preliminaries}
\label{preliminaries}

\textbf{Stable Diffusion Model.} In a T2I task, a generative model receives a text prompt as input and generates an image that precisely matches the prompt. One of the most widely used models in this task is the Stable Diffusion model (SD) \cite{rombach2022high}. SD belongs to the family of latent diffusion models and therefore operates within the latent space of an autoencoder. Denoising Diffusion Probabilistic Model (DDPM) \cite{ho2020denoising} runs with a UNet \cite{ronneberger2015u} backbone (for both SD1 and SD2) in this latent space. In the denoising process, SD gradually denoises the initial noise sampled from a standard normal distribution. If the initial noise differs, the generated image is also different. 

For T2I tasks, SD incorporates cross-attention layers. The text encoder of SD transforms the input text prompt into a sequence of embeddings, which are then projected to serve as keys, while the UNet’s intermediate features undergo a similar projection to serve as queries. In addition, SD includes self-attention layers, in which the UNet’s intermediate features serve as both keys and queries. The resulting cross-attention maps and self-attention maps derived from these key-query pairs are used by all the algorithms examined in this paper.

\section{Best-of-N Inference-Time Scaling}

\subsection{Metrics for the given initial noise}

In this study, we apply Best-of-N inference-time scaling to the T2I diffusion model by sampling multiple initial noises and selecting the best one. This is similar to one of the methods of the previous study \cite{ma2025inference} which applied inference-time scaling to the T2I diffusion model with an additional verifier model. However, we evaluate the quality of a given initial noise with loss functions of various algorithms that optimize the initial noise of diffusion models. CONFORM \cite{meral2024conform} utilizes the cross-attention maps of T2I diffusion models for contrastive learning. It employs InfoNCE \cite{oord2018representation} as its loss, grouping the objects and attributes within the input text prompt into positive and negative pairs. InitNO \cite{guo2024initno} uses the sum of two metrics as its loss. First, it considers $1 - \emph{\textbf{minmax\_cross}}$, where $\emph{\textbf{minmax\_cross}}$ is the smallest value among the maximum cross-attention weight values for each object that must appear in the resulting image. Second, InitNO measures the degree of overlap between self-attention maps corresponding to the spatial patch with the largest cross-attention weight of each object. Although the paper formally includes a regularization term to prevent deviations from the standard normal distribution for every optimization step, the official implementation\footnote{\texttt{https://github.com/xiefan-guo/initno/tree/main}} uses this term in optimization process only when it exceeds a certain threshold. Since we sample noises from a standard normal distribution, we do not consider this regularization term. The loss of Self-Cross guidance \cite{qiu2024self} is largely similar to that of InitNO. The key difference is that it does not focus solely on the self-attention map of the patch with the largest cross-attention weight; instead, it uses the entire patches' self-attention maps with each of their cross-attention weights when computing overlaps between objects. The authors propose to use InitNO before Self-Cross guidance.

\subsection{Best-of-N}

To examine whether allocating more compute is beneficial for initial noise optimization algorithms that rely solely on a T2I diffusion model, we adopt a Best-of-N approach to the number of candidate initial noises in each algorithm. We allocate a total of $N$ loss calculations, assess each candidate noise using these loss values, and select the noise achieving the best (lowest) loss. Note that unlike the other two algorithms, InitNO \cite{guo2024initno} optimizes a single initial noise multiple times. Concretely, InitNO first samples an initial noise and runs up to 10 optimization steps, checking at each step whether the resulting noise is valid, i.e., the loss value falls under the predefined threshold. If the noise becomes valid at any step, the optimization is halted immediately and that noise is used as the initial noise for the diffusion model. If it remains invalid after all 10 steps, a new noise is sampled, and the process is repeated up to 5 times. If all 5 trials are invalid, InitNO picks the noise that has the smallest loss among them, optimizes it for more 40 iterations, and ultimately uses that result as the initial noise for the diffusion model. As a result, in the case of InitNO, each noise candidate is optimized over 10 loss calculations (without early stopping for fairness), so the effective number of candidates is $\tfrac{N}{10}$. In contrast, both CONFORM \cite{meral2024conform} and Self-Cross guidance \cite{qiu2024self} retain $N$ candidates. As $N$ increases, the number of candidate initial noises will also increase. If the loss function of each algorithm accurately reflects the quality of the resulting image, the performance will always increase monotonically as the candidate initial noise increases. And if this is true, we can conclude that applying inference-time scaling to algorithms that only use the soley T2I diffusion model helps improve performance. We conduct extensive experiments to confirm this.

\begin{figure}[t]
\vskip 0.1in
\begin{center}
\centerline{\includegraphics[width=0.8\columnwidth]{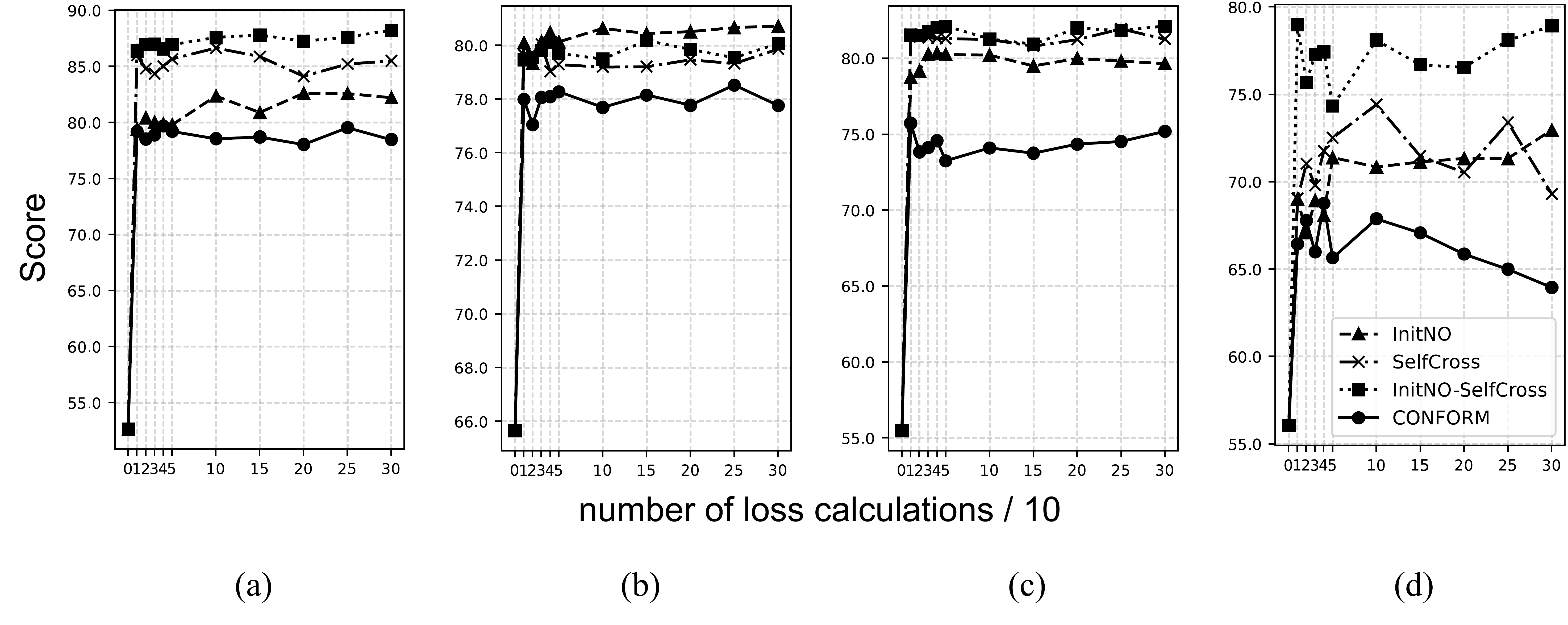}}
\caption{\textbf{Experimental results in SD1.5.} We conducted experiments on four datasets in SD1.5: (a) \emph{animal\_animal}, (b) \emph{animal\_object}, (c) \emph{object\_object}, (d) \emph{similar\_subjects}. In each graph, the horizontal axis (x-axis) represents the number of loss calculations $N$ divided by 10 ($N / 10$), while the vertical axis (y-axis) shows the resulting scores. The legend in (d) lists the names of each algorithm evaluated.}
\label{SD15 result}
\end{center}
\vskip -0.05in
\end{figure}

\section{Experiments}

\subsection{Experimental settings}

We followed the experimental settings described in the Self-Cross guidance paper \cite{qiu2024self}. The input text prompts are categorized into four datasets: (1) \emph{animal\_animal}, consisting of 66 prompts featuring various pairs of animals; (2) \emph{animal\_object}, consisting of 144 prompts describing diverse animal–object pairs; (3) \emph{object\_object}, consisting of 66 prompts containing two distinct subjects; and (4) \emph{similar\_subjects}, consisting of 31 prompts illustrating two similar subjects. For the score calculation, we use GPT4o \cite{achiam2023gpt} and predefined questions to determine whether each resulting image (i) includes both subjects (\emph{Existence}), (ii) presents both subjects in a recognizable form (\emph{Recognizability}), and (iii) does not exhibit any undesired mixture between the two subjects (\emph{Not a Mixture}). The percentage of the positive responses from GPT4o was calculated as a score. All experiments were repeated with 10 different random seeds to ensure reliability and statistical robustness of the reported results. In our experiments, we used the official implementation of Self-Cross guidance\footnote{\texttt{https://github.com/mengtang-lab/selfcross-guidance/tree/main}}, which also includes implementations of InitNO \cite{guo2024initno} and CONFORM \cite{meral2024conform}. Since both the InitNO's and Self-Cross guidance's official implementations note that InitNO does not work well on SD2.1 and we also confirmed this in our own tests, we did not use InitNO for SD2.1. Additionally, to investigate how much InitNO contributes to the performance of Self-Cross guidance, we distinguish between the version that uses InitNO first (InitNO-SelfCross) and the one that does not (SelfCross).

\subsection{Results}

\begin{figure}[]
\vskip 0.05in
\begin{center}
\centerline{\includegraphics[width=0.8\columnwidth]{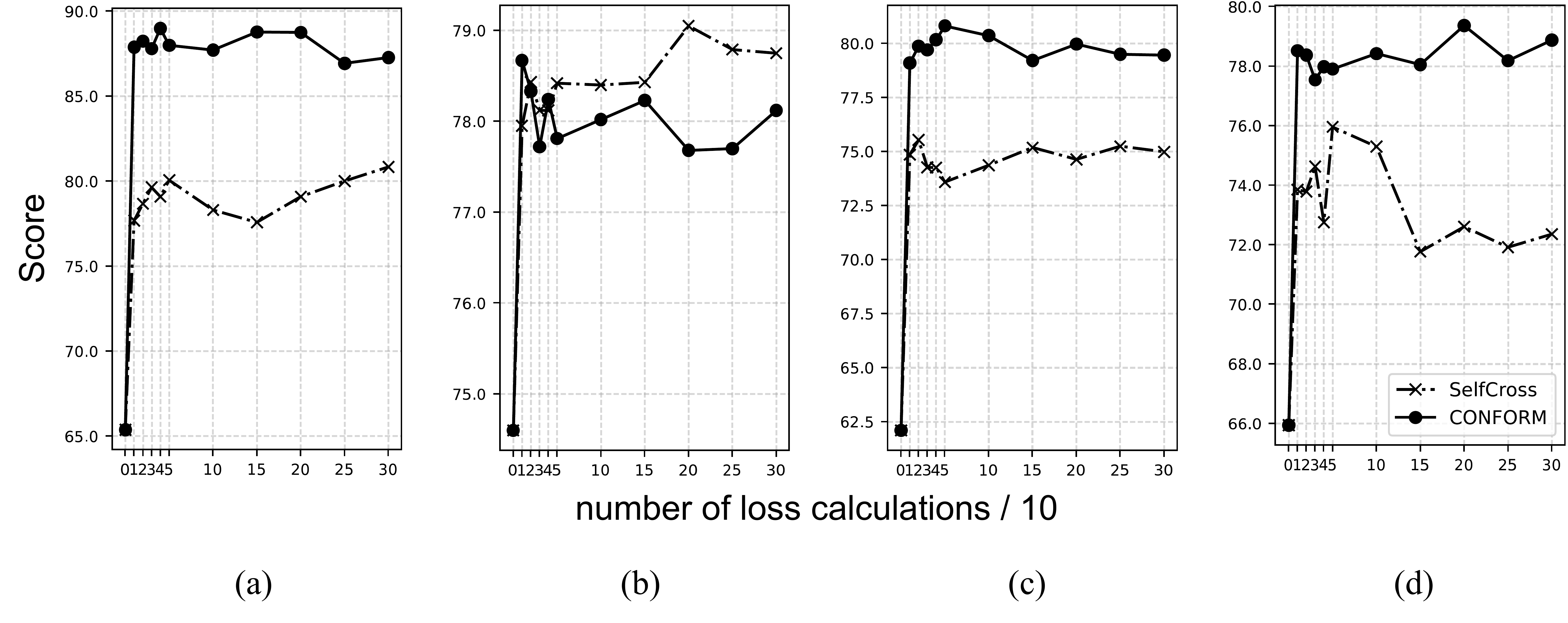}}
\caption{\textbf{Experimental results in SD2.1.} We conducted experiments on four datasets in SD2.1: (a) \emph{animal\_animal}, (b) \emph{animal\_object}, (c) \emph{object\_object}, (d) \emph{similar\_subjects}. In each graph, the horizontal axis (x-axis) represents the number of loss calculations $N$ divided by 10 ($N / 10$), while the vertical axis (y-axis) shows the resulting scores. The legend in (d) lists the names of each algorithm evaluated.}
\label{SD21 result}
\vskip -0.1in
\end{center}
\end{figure}

Figure \ref{SD15 result} and \ref{SD21 result} present the main experimental results, while more detailed quantitative and qualitative findings are provided in the Appendix. When the number of loss calculations $N$ is set to 0, the result corresponds to the default SD's output without any additional algorithms. For SD1.5, InitNO-SelfCross \cite{qiu2024self} performs best overall, as this is the last proposed algorithm. For SD2.1, CONFORM \cite{meral2024conform} is generally the top-performing method in most cases, even though this is the earliest algorithm among the algorithms used in this study.

As we increase the number of candidate initial noises, we can see the result that contradicts the common expectation. \textbf{In multiple backbones, various algorithms do not guarantee an increase in performance as candidate initial noise increases}. Although the default setting for InitNO \cite{guo2024initno} is $N = 50$ and the general expectation is that larger $N$ should offer better results, all initial noise optimization algorithms do not show a consistent performance increase beyond $N = 10$. In other words, for InitNO, using just a single noise sample and optimizing it suffices to achieve the maximum possible performance with the algorithm and the other two algorithms also show similar patterns. 

\begin{figure}[]
\vskip 0.1in
\begin{center}
\centerline{\includegraphics[width=0.4\columnwidth]{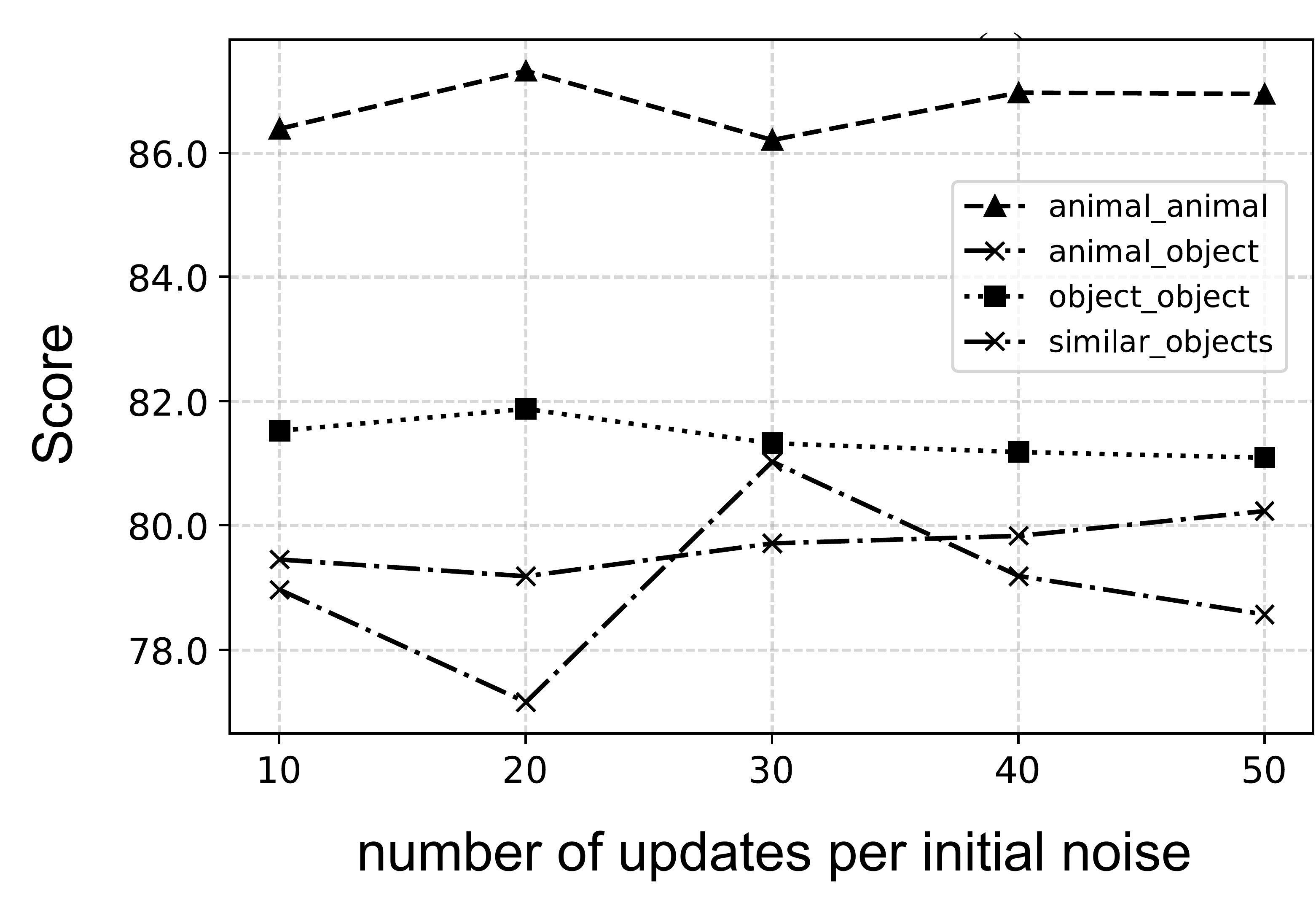}}
\caption{\textbf{Experimental results in InitNO.} We conducted experiments on four datasets in SD2.1. In each graph, the horizontal axis (x-axis) represents the number of updates per initial noise, while the vertical axis (y-axis) shows the resulting scores. The legend lists the names of the prompt datasets.}
\label{InitNO Result}
\vskip -0.1in
\end{center}
\end{figure}

These findings suggest that \textbf{the losses employed by current initial noise optimization algorithms do not perfectly capture the alignment between the input text prompt and the resulting image}. To examine this point more closely, we varied the number of optimization iterations in the InitNO phase that is conducted first in the Self-Cross guidance. We retained the original setting of 5 candidate initial noise samples but increased the optimization steps for each sample from 10 to 50 gradually. As Figure \ref{InitNO Result} shows, more optimization steps for the initial noise is not helpful for the performance. 

We confirmed that the performance does not increase even if we sample more random noises and select the noise with the smallest loss among them, or further optimize using the loss of the sampled noise. Nevertheless, the performance clearly increased compared to not using any of these algorithms. Therefore, we find that it is better to optimize the initial noise with the least computational resource (but not zero) for GPUs with limited VRAM.

\section{Conclusion}

In this study, we found that when we only have VRAM-limited GPUs, the most efficient way to improve the performance of the T2I diffusion model is to optimize the initial noise with the least computational resources. In addition, we can see that existing training-free approaches for initial noise optimization have considerable potential for further improvement.

\section*{Acknowledgements}

This work was supported by Institute of Information \& communications Technology Planning \& Evaluation (IITP) grant funded by the Korea government(MSIT) [NO.RS-2021-II211343, Artificial Intelligence Graduate School Program (Seoul National University)]

\bibliography{main}
\bibliographystyle{icml2025}

\clearpage

\section*{Appendix}
\label{appendix}

Table \ref{appendix table 1}, \ref{appendix table 2}, \ref{appendix table 3} and \ref{appendix table 4} present the quantitative results corresponding to Figure \ref{SD15 result} and \ref{SD21 result}. Table \ref{appendix figure 1} shows the qualitative outcomes for the input text prompt “\emph{a bear and a red balloon}”.

\vskip 0.2in

\FloatBarrier

\begin{table}[ht]
\caption{\textbf{Quantitative results for each initial noise optimization algorithm for \emph{animal\_animal} dataset.} $N$ represents the number of loss calculations. The numbers in the parentheses indicate the versions of Stable Diffusion models.}
\label{appendix table 1}
\vskip 0.1in
\begin{center}
\begin{tabular}{l|cccc|cr}
\toprule
$N$ & CONFORM(1.5) & InitNO(1.5) & SelfCross(1.5) & InitNO-SelfCross(1.5) & CONFORM(2.1) & SelfCross(2.1) \\
\midrule
0           & 52.64 & 52.64 & 52.64 & 52.64 & 65.38 & 65.38 \\
10          & 79.22 & 79.41 & 86.00 & 86.39 & 87.88 & 77.68 \\
20          & 78.53 & 80.41 & 84.81 & 86.95 & 88.23 & 78.67 \\
30          & 78.89 & 80.02 & 84.35 & 86.99 & 87.79 & 79.64 \\
40          & 79.69 & 79.88 & 85.04 & 86.58 & 88.98 & 79.09 \\
50          & 79.22 & 79.79 & 85.67 & 86.93 & 87.99 & 80.06 \\
100         & 78.56 & 82.36 & 86.65 & 87.58 & 87.71 & 78.31 \\
150         & 78.72 & 80.89 & 85.89 & 87.80 & 88.77 & 77.58 \\
200         & 78.05 & 82.60 & 84.13 & 87.25 & 88.75 & 79.09 \\
250         & 79.55 & 82.58 & 85.22 & 87.60 & 86.93 & 80.01 \\
300         & 78.48 & 82.21 & 85.50 & 88.23 & 87.27 & 80.84 \\
\bottomrule
\end{tabular}
\end{center}
\vskip -0.1in
\end{table}

\vskip 1.5in

\begin{table}[ht]
\caption{\textbf{Quantitative results for each initial noise optimization algorithm for \emph{animal\_object} dataset.} $N$ represents the number of loss calculations. The numbers in the parentheses indicate the versions of Stable Diffusion models.}
\label{appendix table 2}
\vskip 0.1in
\begin{center}
\begin{tabular}{l|cccc|cr}
\toprule
$N$ & CONFORM(1.5) & InitNO(1.5) & SelfCross(1.5) & InitNO-SelfCross(1.5) & CONFORM(2.1) & SelfCross(2.1) \\
\midrule
0           & 65.66 & 65.66 & 65.66 & 65.66 & 74.60 & 74.60 \\
10          & 77.99 & 80.10 & 79.65 & 79.46 & 78.67 & 77.95 \\
20          & 77.05 & 79.34 & 79.71 & 79.54 & 78.33 & 78.43 \\
30          & 78.06 & 80.14 & 80.05 & 79.81 & 77.72 & 78.12 \\
40          & 78.09 & 80.48 & 79.04 & 80.11 & 78.24 & 78.12 \\
50          & 78.27 & 80.13 & 79.29 & 79.70 & 77.81 & 78.42 \\
100         & 77.69 & 80.63 & 79.20 & 79.49 & 78.02 & 78.40 \\
150         & 78.15 & 80.44 & 79.21 & 80.18 & 78.23 & 78.43 \\
200         & 77.77 & 80.51 & 79.47 & 79.85 & 77.68 & 79.05 \\
250         & 78.52 & 80.66 & 79.33 & 79.54 & 77.70 & 78.79 \\
300         & 77.76 & 80.72 & 79.87 & 80.07 & 78.12 & 78.75 \\
\bottomrule
\end{tabular}
\end{center}
\vskip -0.1in
\end{table}

\begin{table}[ht]
\caption{\textbf{Quantitative results for each initial noise optimization algorithm for \emph{object\_object} dataset.} $N$ represents the number of loss calculations. The numbers in the parentheses indicate the versions of Stable Diffusion models.}
\label{appendix table 3}
\vskip 0.1in
\begin{center}
\begin{tabular}{l|cccc|cr}
\toprule
$N$ & CONFORM(1.5) & InitNO(1.5) & SelfCross(1.5) & InitNO-SelfCross(1.5) & CONFORM(2.1) & SelfCross(2.1) \\
\midrule
0           & 55.50 & 55.50 & 55.50 & 55.50 & 62.10 & 62.10 \\
10          & 75.72 & 78.72 & 81.58 & 81.53 & 79.09 & 74.85 \\
20          & 73.84 & 79.17 & 81.40 & 81.49 & 79.87 & 75.53 \\
30          & 74.13 & 80.28 & 81.37 & 81.74 & 79.70 & 74.26 \\
40          & 74.57 & 80.35 & 81.32 & 82.04 & 80.17 & 74.25 \\
50          & 73.25 & 80.25 & 81.31 & 82.12 & 80.81 & 73.58 \\
100         & 74.10 & 80.21 & 81.25 & 81.27 & 80.36 & 74.36 \\
150         & 73.75 & 79.50 & 80.81 & 80.92 & 79.21 & 75.18 \\
200         & 74.35 & 79.99 & 81.24 & 81.99 & 79.97 & 74.63 \\
250         & 74.52 & 79.82 & 81.95 & 81.81 & 79.50 & 75.24 \\
300         & 75.19 & 79.65 & 81.27 & 82.12 & 79.46 & 74.98 \\
\bottomrule
\end{tabular}
\end{center}
\vskip -0.1in
\end{table}

\begin{table}[ht]
\caption{\textbf{Quantitative results for each initial noise optimization algorithm for \emph{similar\_subjects} dataset.} $N$ represents the number of loss calculations. The numbers in the parentheses indicate the versions of Stable Diffusion models.}
\label{appendix table 4}
\vskip 0.1in
\begin{center}
\begin{tabular}{l|cccc|cr}
\toprule
$N$ & CONFORM(1.5) & InitNO(1.5) & SelfCross(1.5) & InitNO-SelfCross(1.5) & CONFORM(2.1) & SelfCross(2.1) \\
\midrule
0           & 56.06 & 56.06 & 56.06 & 56.06 & 65.94 & 65.94 \\
10          & 66.44 & 69.00 & 69.06 & 78.97 & 78.52 & 73.85 \\
20          & 67.78 & 67.09 & 71.03 & 75.69 & 78.37 & 73.79 \\
30          & 65.99 & 68.92 & 69.80 & 77.28 & 77.54 & 74.63 \\
40          & 68.77 & 68.08 & 71.77 & 77.44 & 77.98 & 72.76 \\
50          & 65.65 & 71.38 & 72.52 & 74.33 & 77.90 & 75.96 \\
100         & 67.89 & 70.84 & 74.43 & 78.10 & 78.42 & 75.30 \\
150         & 67.08 & 71.13 & 71.48 & 76.70 & 78.05 & 71.77 \\
200         & 65.87 & 71.33 & 70.54 & 76.55 & 79.36 & 72.61 \\
250         & 65.00 & 71.33 & 73.40 & 78.10 & 78.18 & 71.92 \\
300         & 63.96 & 72.96 & 69.31 & 78.92 & 78.87 & 72.36 \\
\bottomrule
\end{tabular}
\end{center}
\vskip -0.1in
\end{table}

\begin{table}[ht]
\caption{\textbf{Resulting images for each initial noise optimization algorithm.} Input text prompt is “\emph{a bear and a red balloon}”. $N$ represents the number of loss calculations. The numbers in the parentheses indicate the versions of Stable Diffusion models.}
\label{appendix figure 1}
\vskip 0.1in
\begin{center}
\begin{tabular}{l|cccc|cr}
\toprule
$N$ & CONFORM(1.5) & InitNO(1.5) & SelfCross(1.5) & InitNO-SelfCross(1.5) & CONFORM(2.1) & SelfCross(2.1) \\
\midrule
0           & \includegraphics[width=0.1\textwidth, height=17mm]{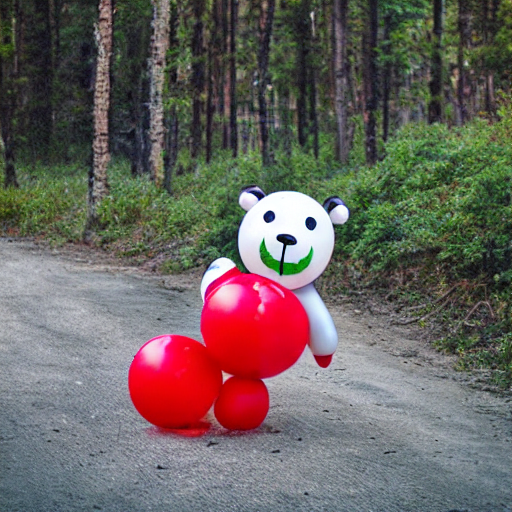} & \includegraphics[width=0.1\textwidth, height=17mm]{fig_result/Original.png} & \includegraphics[width=0.1\textwidth, height=17mm]{fig_result/Original.png} & \includegraphics[width=0.1\textwidth, height=17mm]{fig_result/Original.png} & \includegraphics[width=0.1\textwidth, height=17mm]{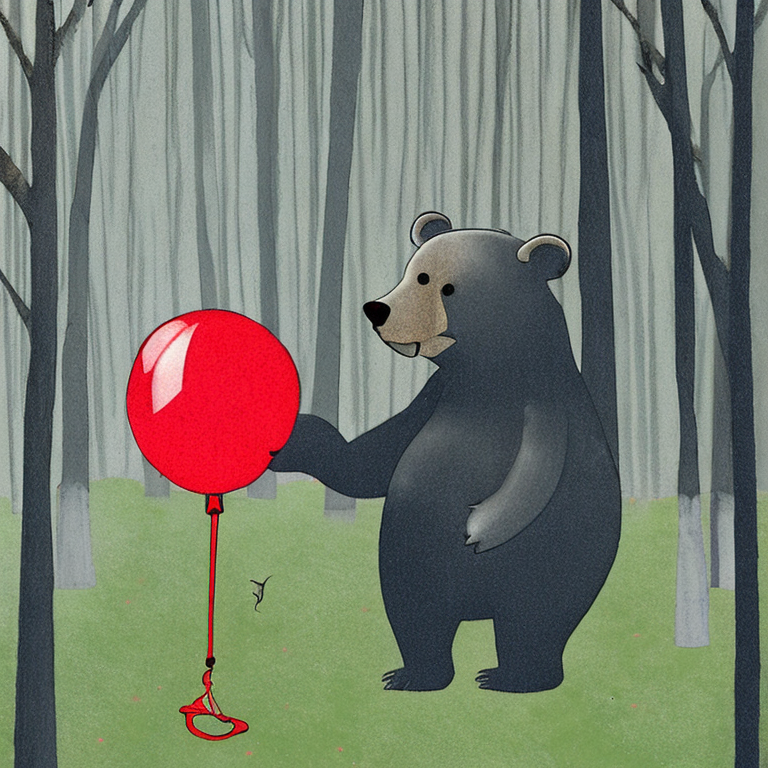} & \includegraphics[width=0.1\textwidth, height=17mm]{fig_result/Original_2.png} \\
10          & \includegraphics[width=0.1\textwidth, height=17mm]{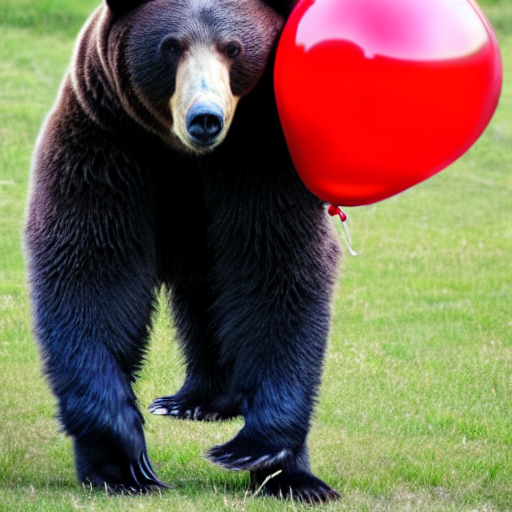} & \includegraphics[width=0.1\textwidth, height=17mm]{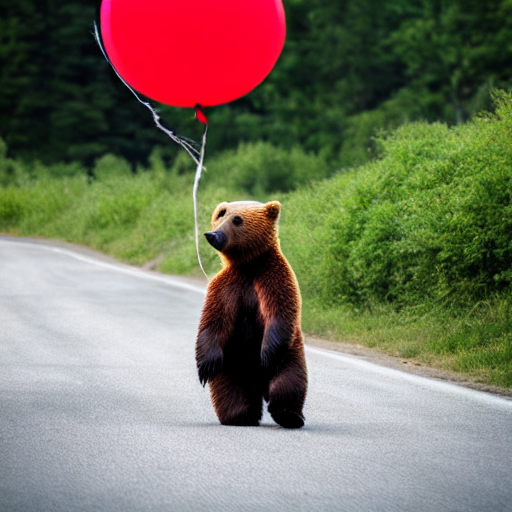} & \includegraphics[width=0.1\textwidth, height=17mm]{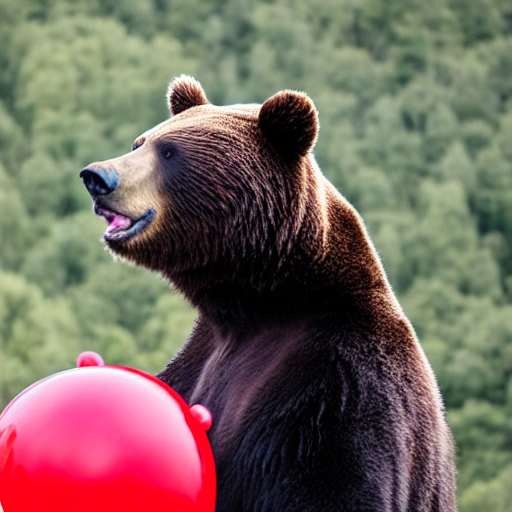} & \includegraphics[width=0.1\textwidth, height=17mm]{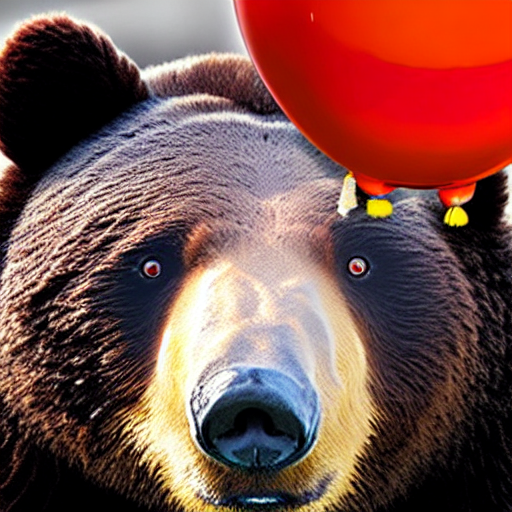} & \includegraphics[width=0.1\textwidth, height=17mm]{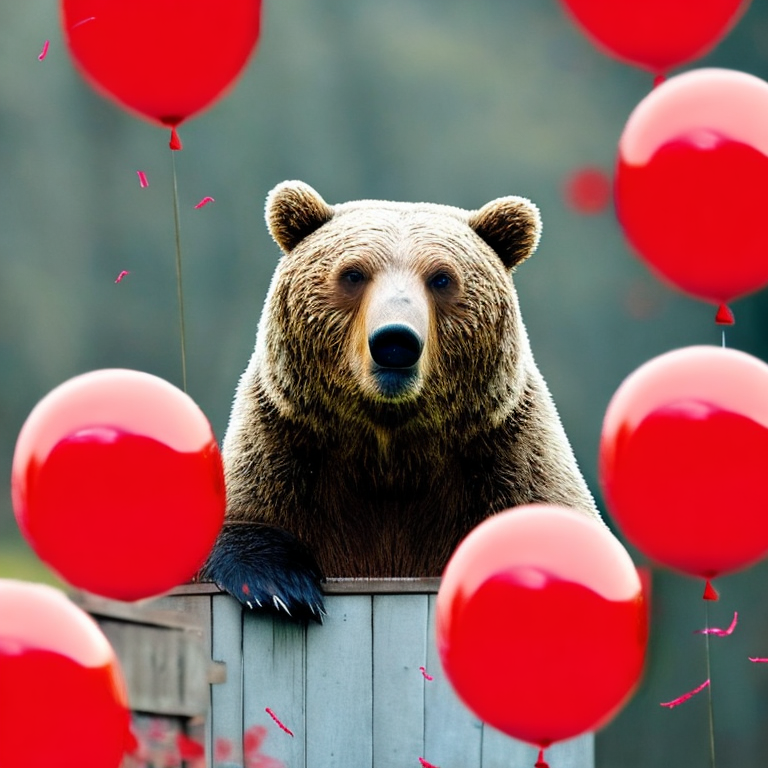} & \includegraphics[width=0.1\textwidth, height=17mm]{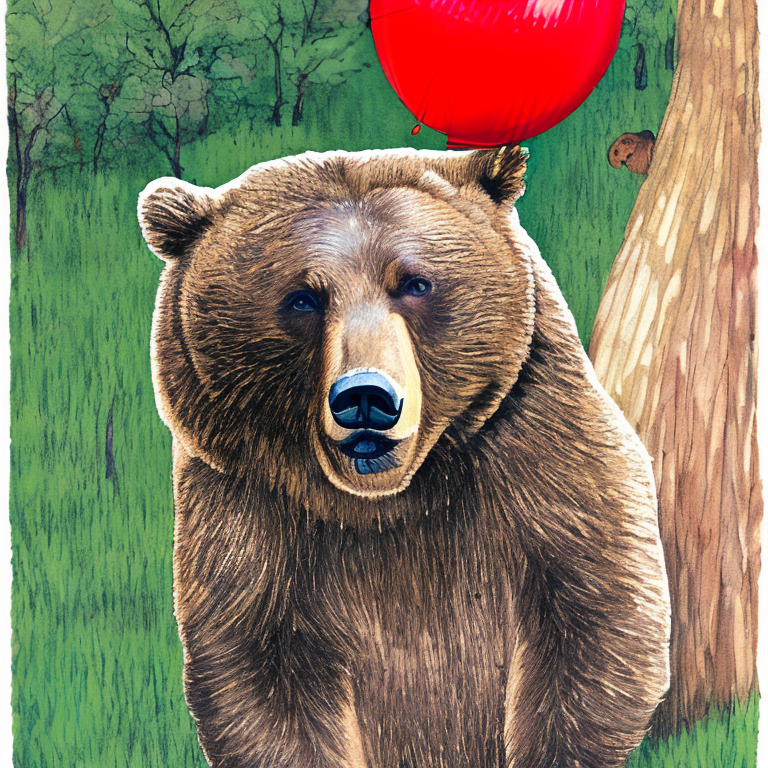} \\
20          & \includegraphics[width=0.1\textwidth, height=17mm]{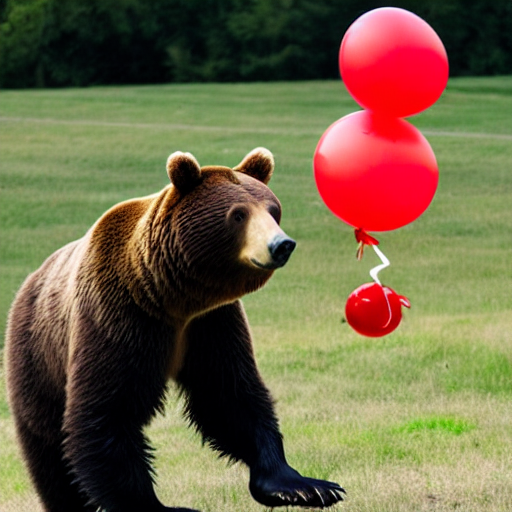} & \includegraphics[width=0.1\textwidth, height=17mm]{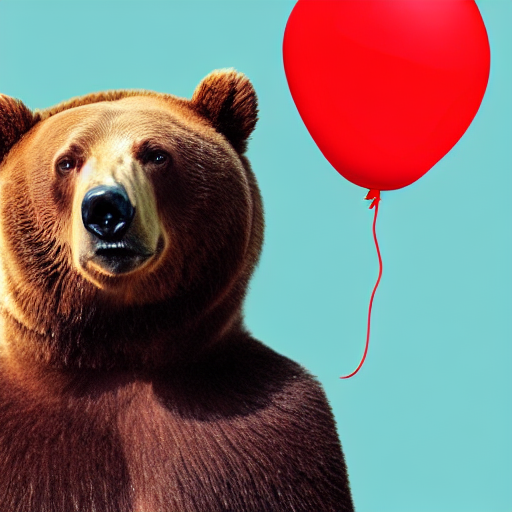} & \includegraphics[width=0.1\textwidth, height=17mm]{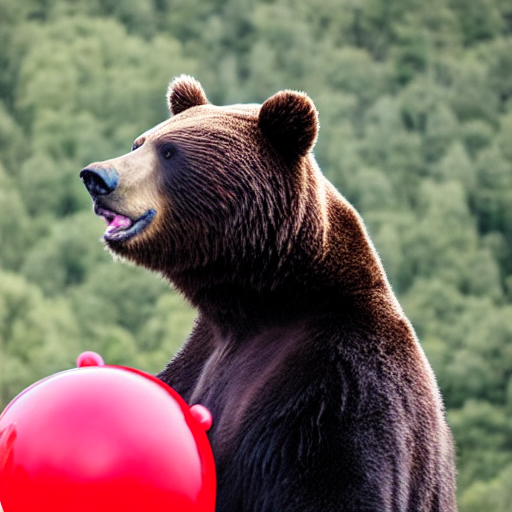} & \includegraphics[width=0.1\textwidth, height=17mm]{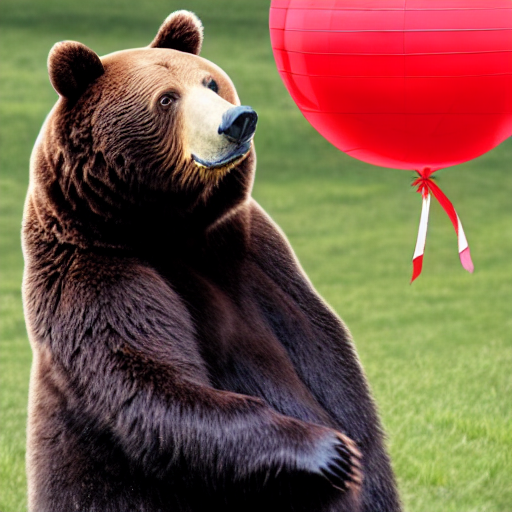} & \includegraphics[width=0.1\textwidth, height=17mm]{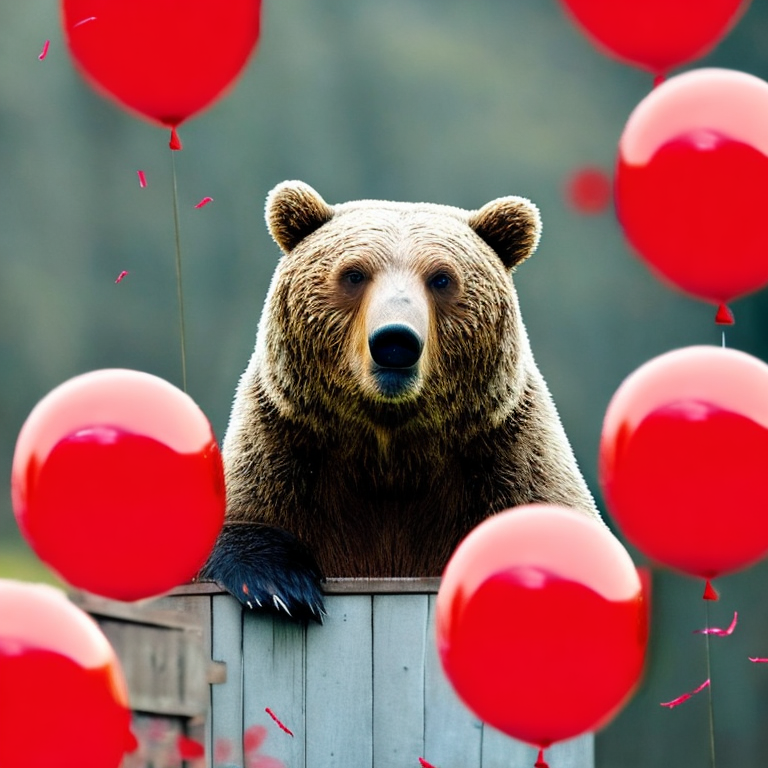} & \includegraphics[width=0.1\textwidth, height=17mm]{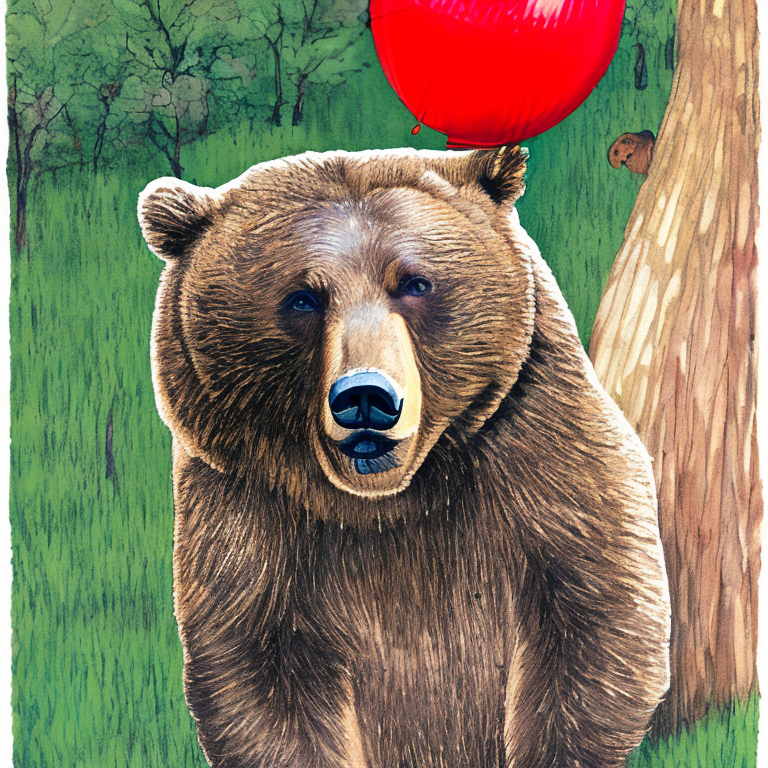} \\
30          & \includegraphics[width=0.1\textwidth, height=17mm]{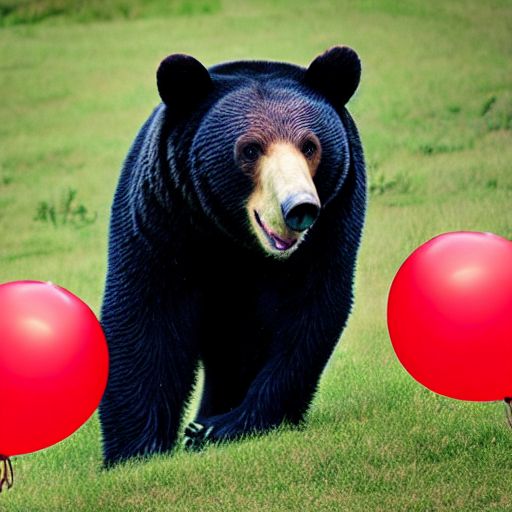} & \includegraphics[width=0.1\textwidth, height=17mm]{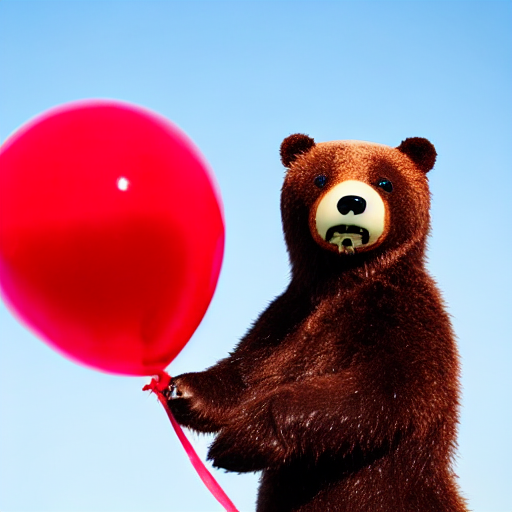} & \includegraphics[width=0.1\textwidth, height=17mm]{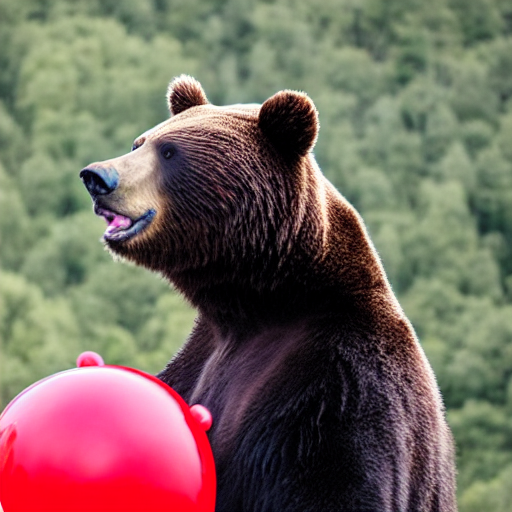} & \includegraphics[width=0.1\textwidth, height=17mm]{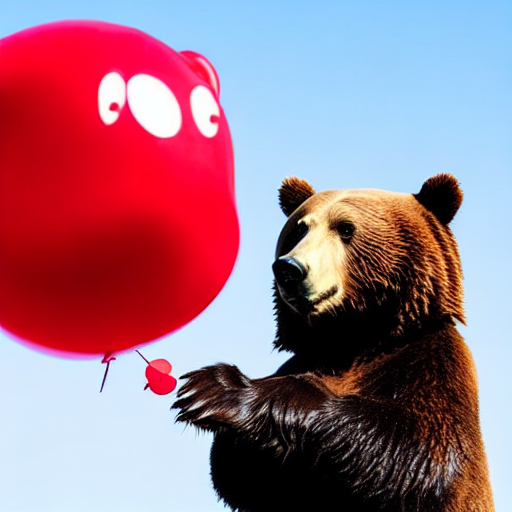} & \includegraphics[width=0.1\textwidth, height=17mm]{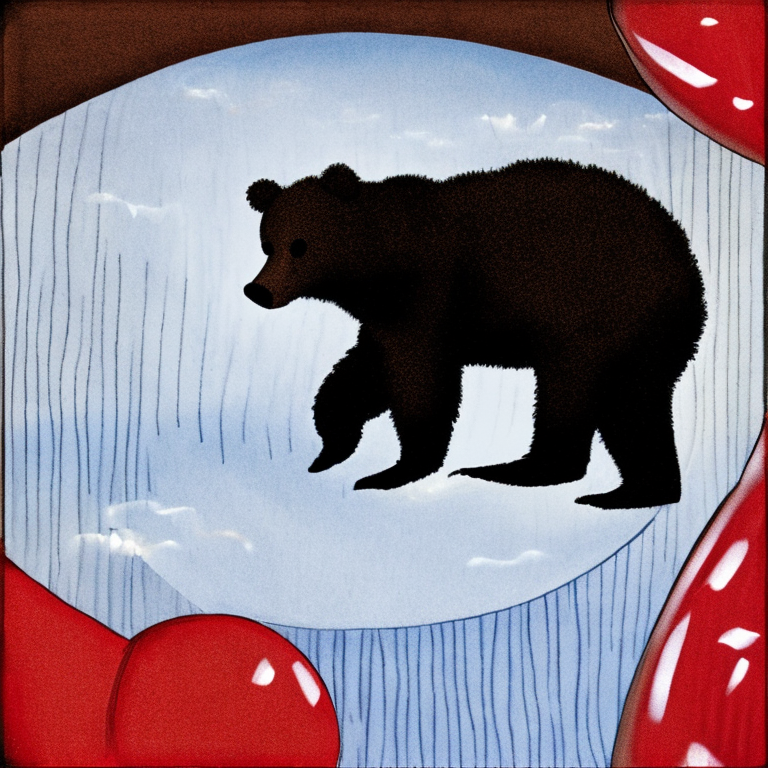} & \includegraphics[width=0.1\textwidth, height=17mm]{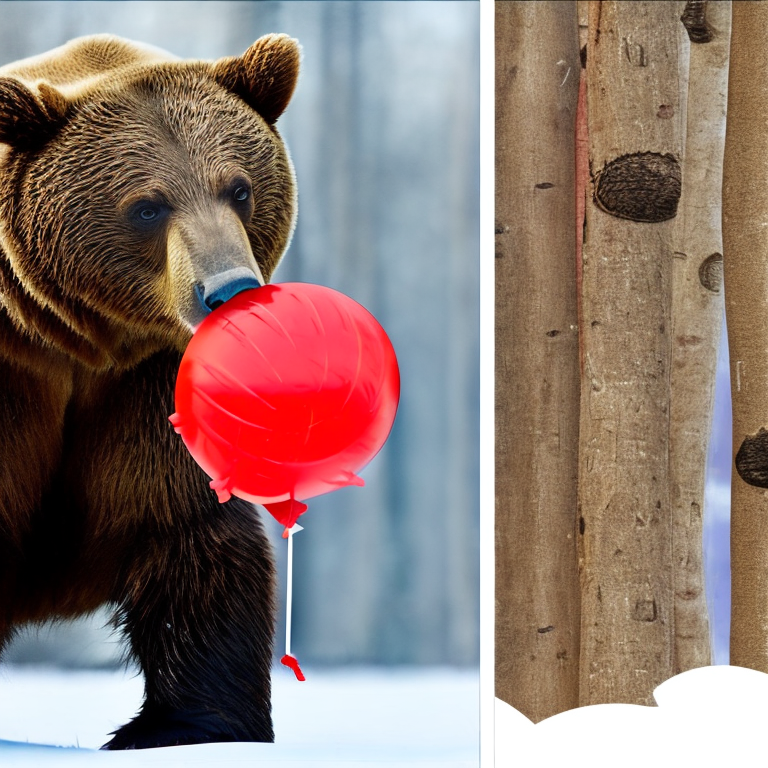} \\
40          & \includegraphics[width=0.1\textwidth, height=17mm]{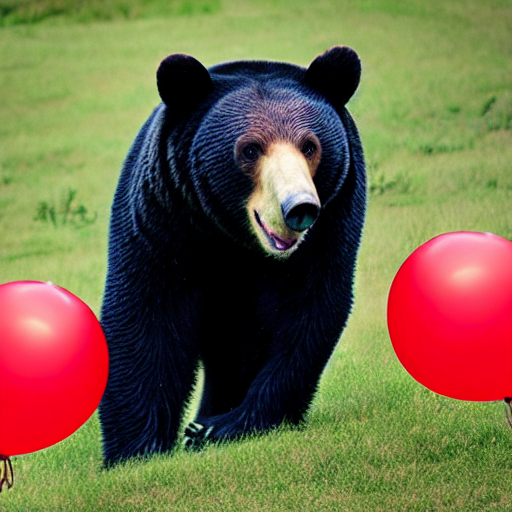} & \includegraphics[width=0.1\textwidth, height=17mm]{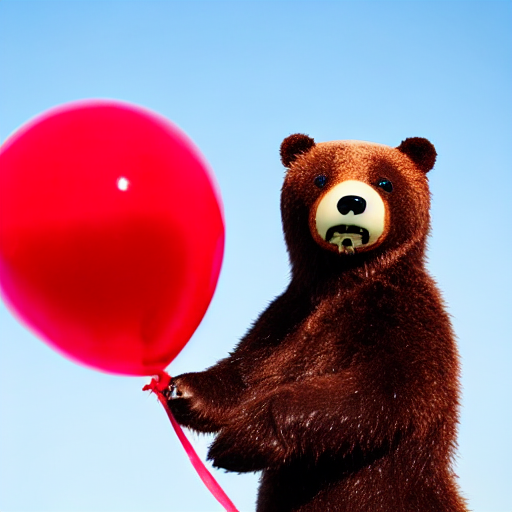} & \includegraphics[width=0.1\textwidth, height=17mm]{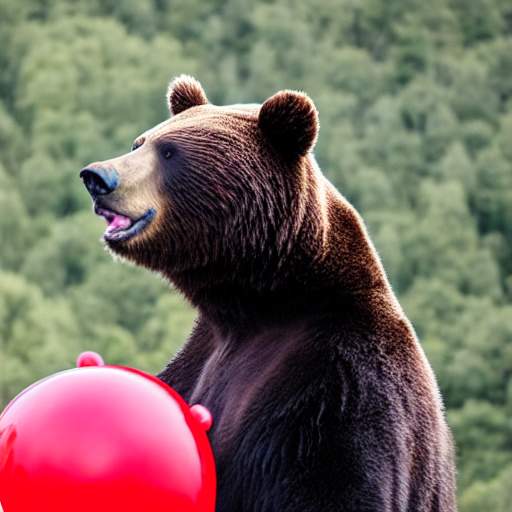} & \includegraphics[width=0.1\textwidth, height=17mm]{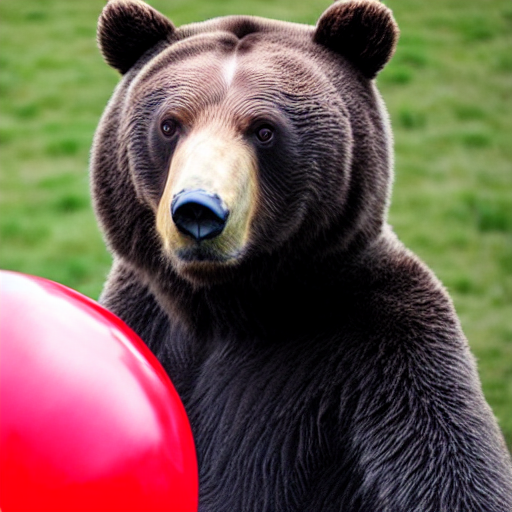} & \includegraphics[width=0.1\textwidth, height=17mm]{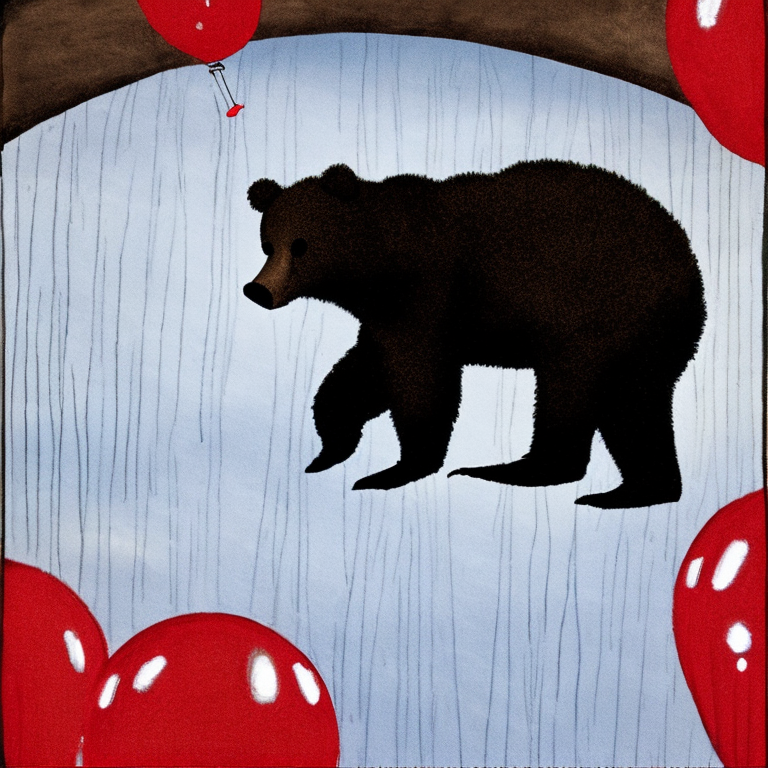} & \includegraphics[width=0.1\textwidth, height=17mm]{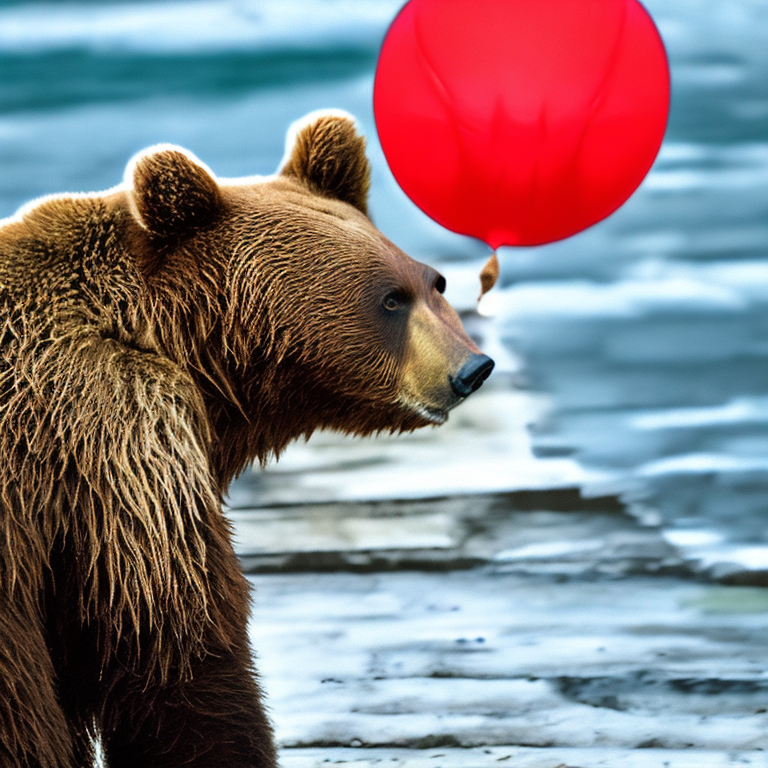} \\
50          & \includegraphics[width=0.1\textwidth, height=17mm]{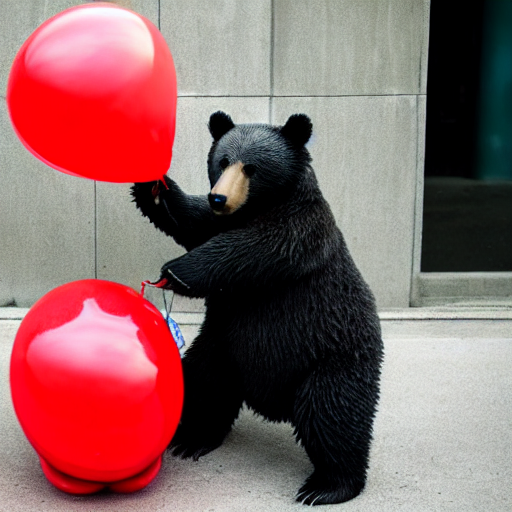} & \includegraphics[width=0.1\textwidth, height=17mm]{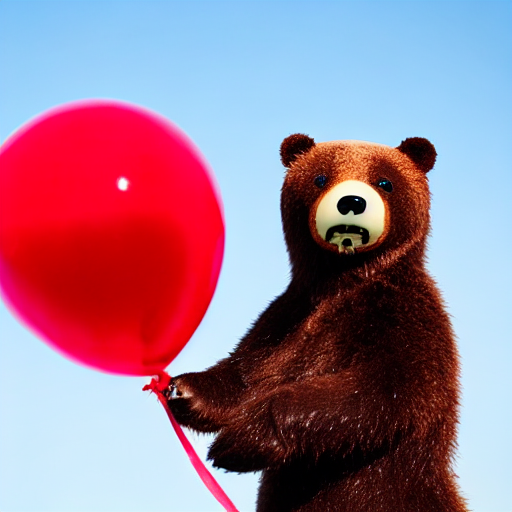} & \includegraphics[width=0.1\textwidth, height=17mm]{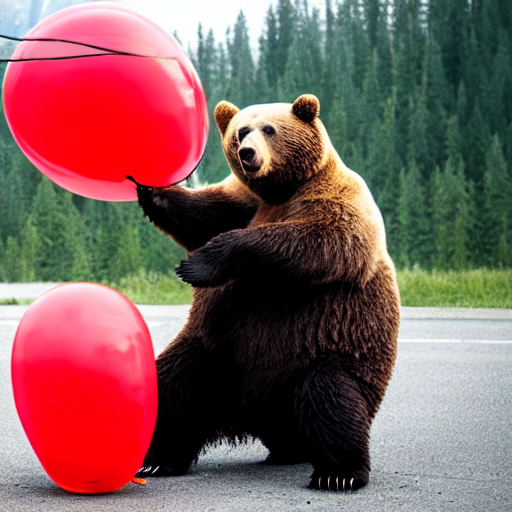} & \includegraphics[width=0.1\textwidth, height=17mm]{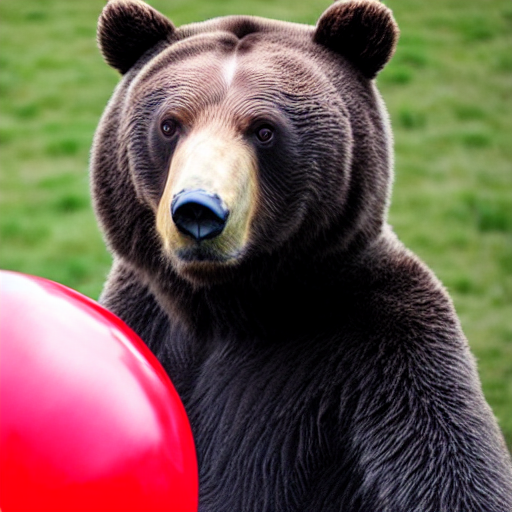} & \includegraphics[width=0.1\textwidth, height=17mm]{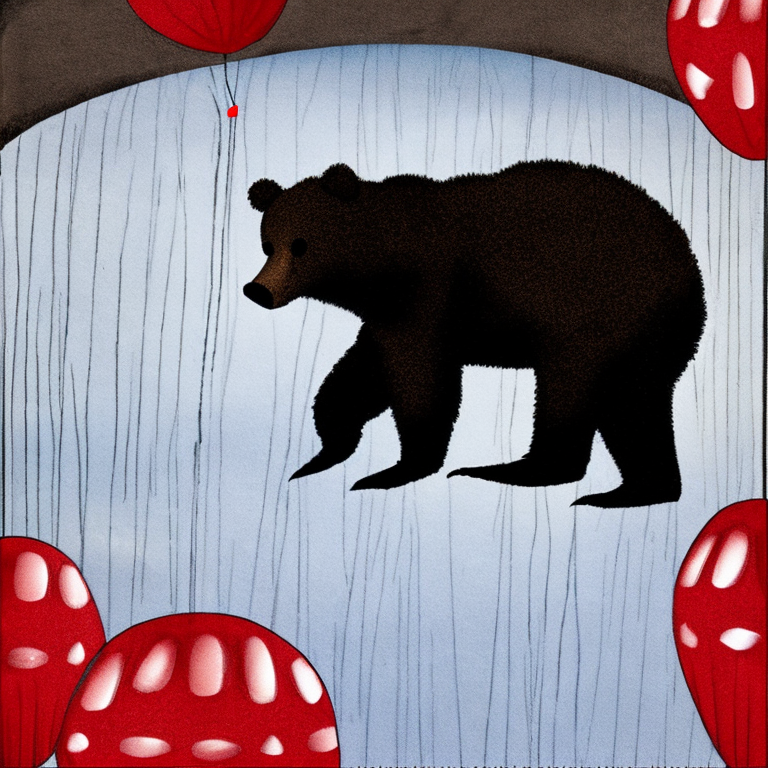} & \includegraphics[width=0.1\textwidth, height=17mm]{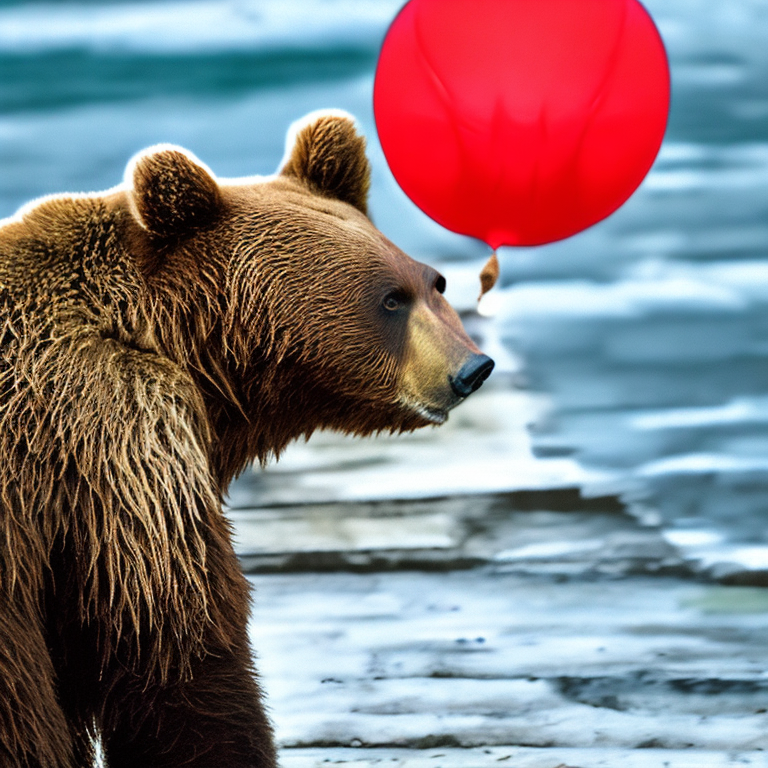} \\
100         & \includegraphics[width=0.1\textwidth, height=17mm]{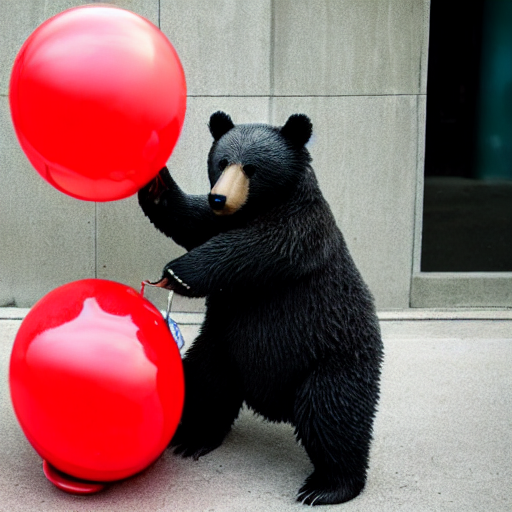} & \includegraphics[width=0.1\textwidth, height=17mm]{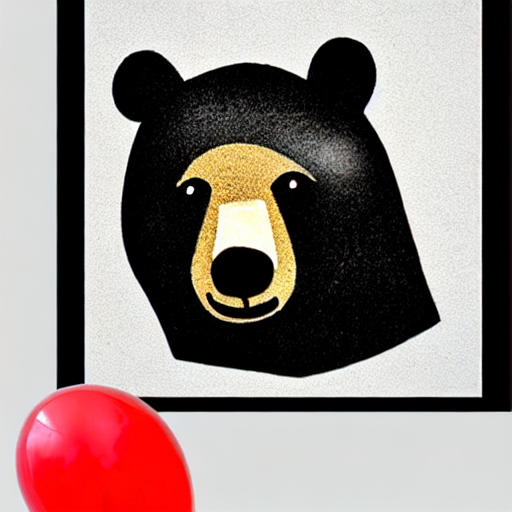} & \includegraphics[width=0.1\textwidth, height=17mm]{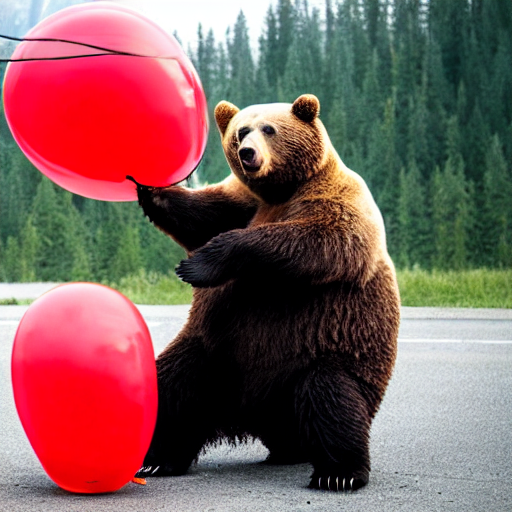} & \includegraphics[width=0.1\textwidth, height=17mm]{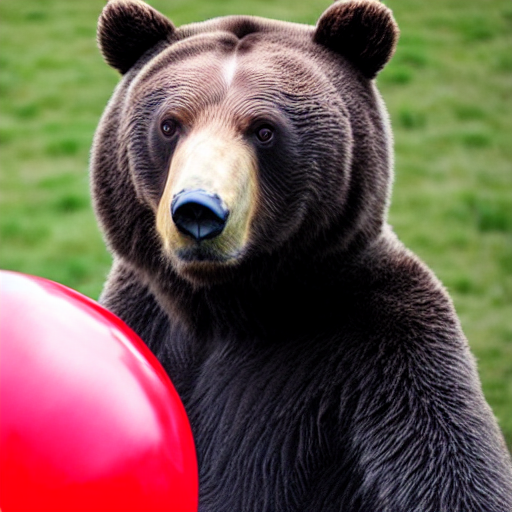} & \includegraphics[width=0.1\textwidth, height=17mm]{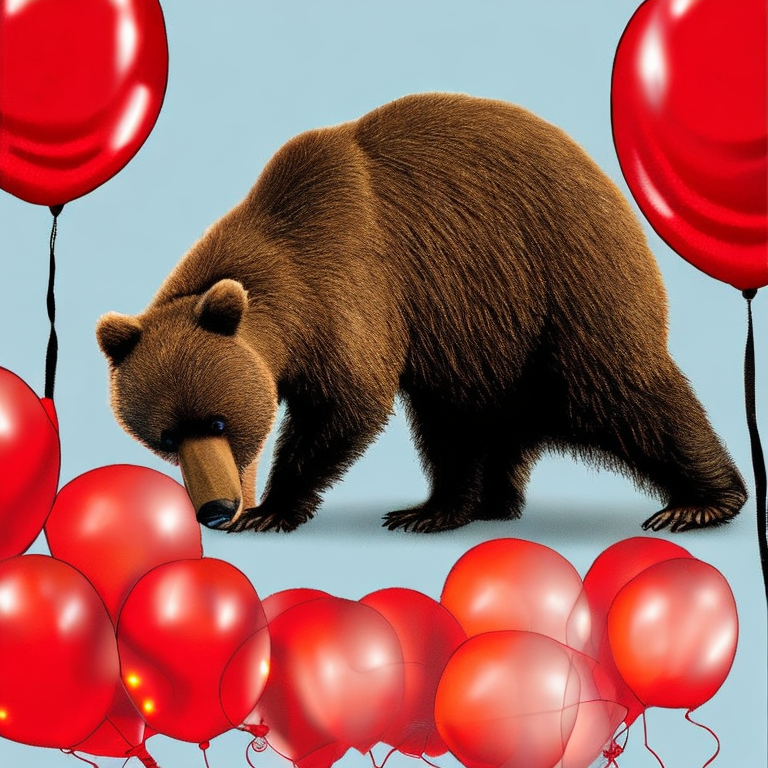} & \includegraphics[width=0.1\textwidth, height=17mm]{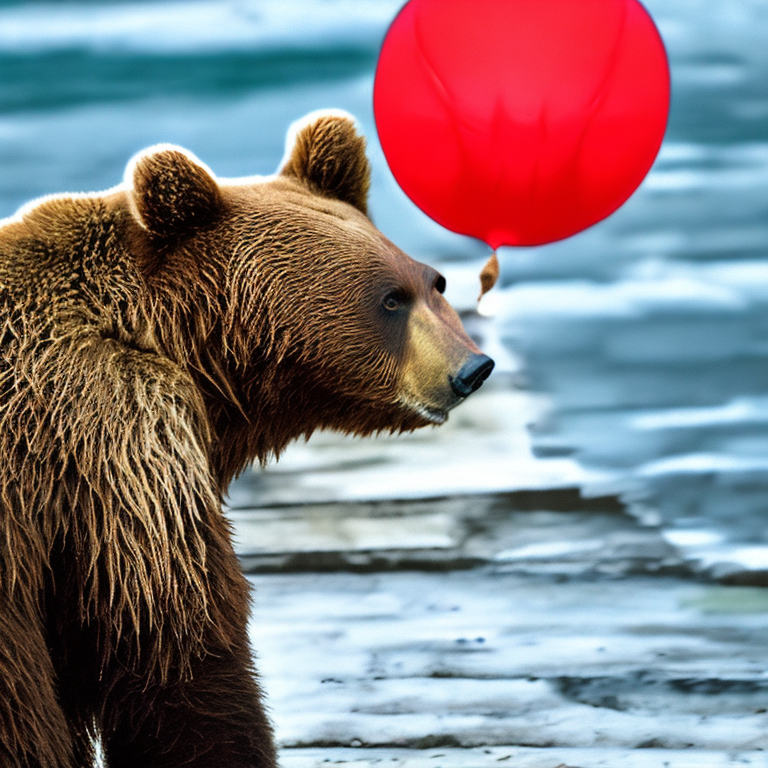} \\
150         & \includegraphics[width=0.1\textwidth, height=17mm]{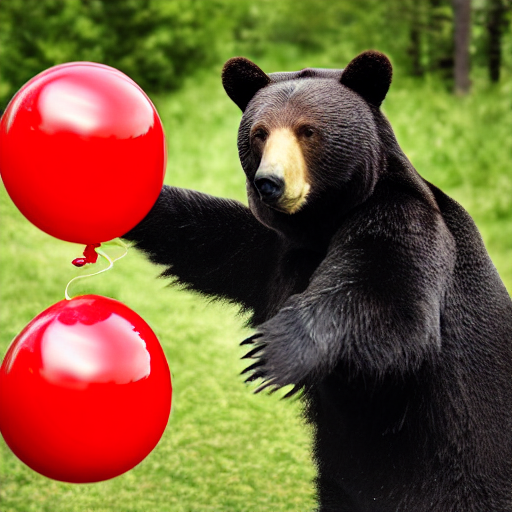} & \includegraphics[width=0.1\textwidth, height=17mm]{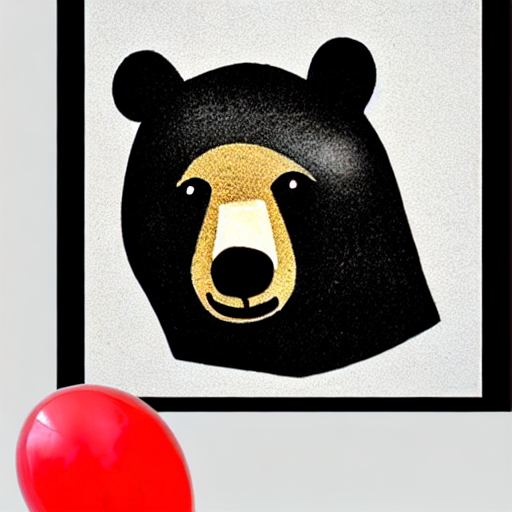} & \includegraphics[width=0.1\textwidth, height=17mm]{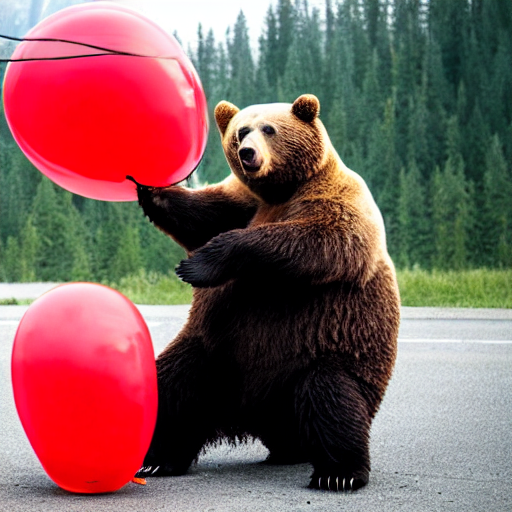} & \includegraphics[width=0.1\textwidth, height=17mm]{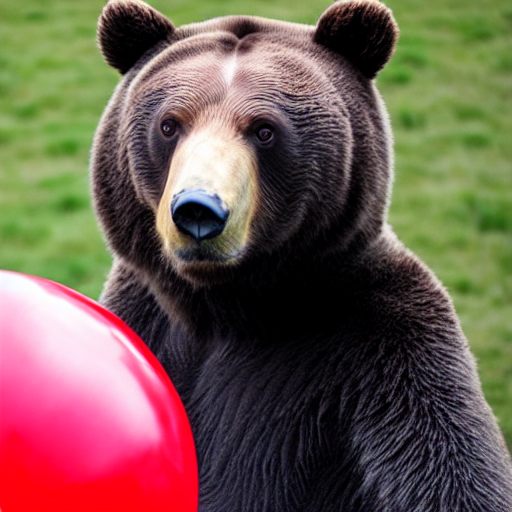} & \includegraphics[width=0.1\textwidth, height=17mm]{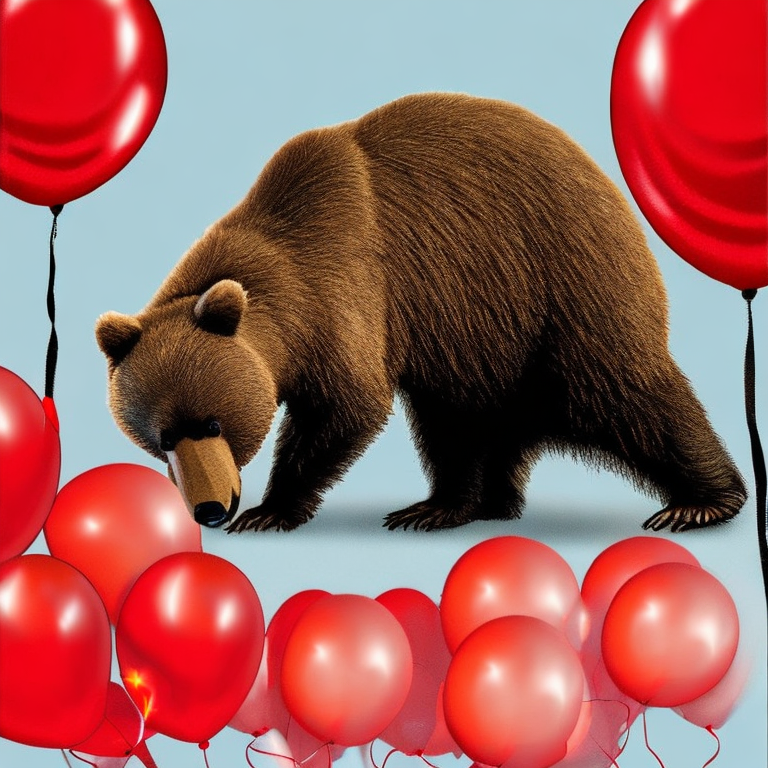} & \includegraphics[width=0.1\textwidth, height=17mm]{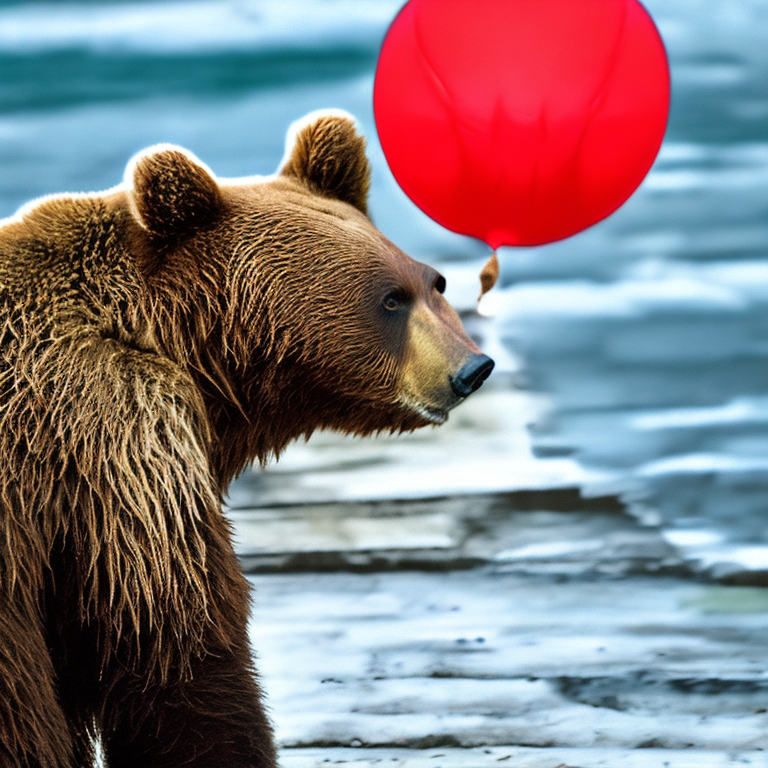} \\
200         & \includegraphics[width=0.1\textwidth, height=17mm]{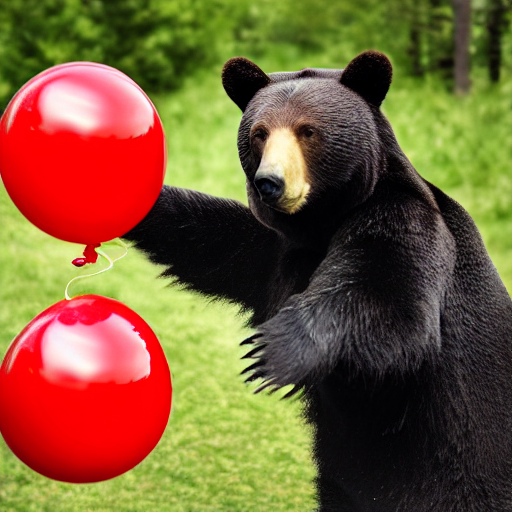} & \includegraphics[width=0.1\textwidth, height=17mm]{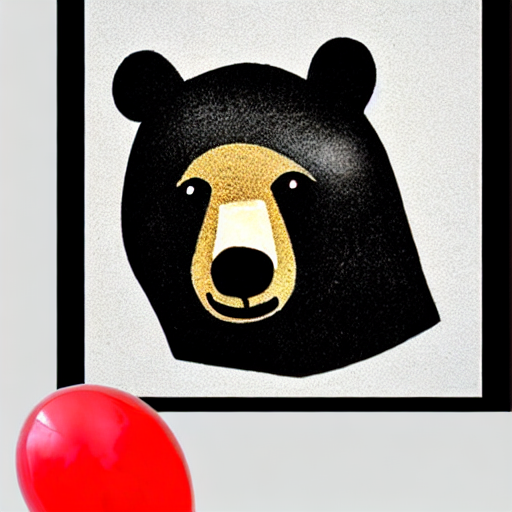} & \includegraphics[width=0.1\textwidth, height=17mm]{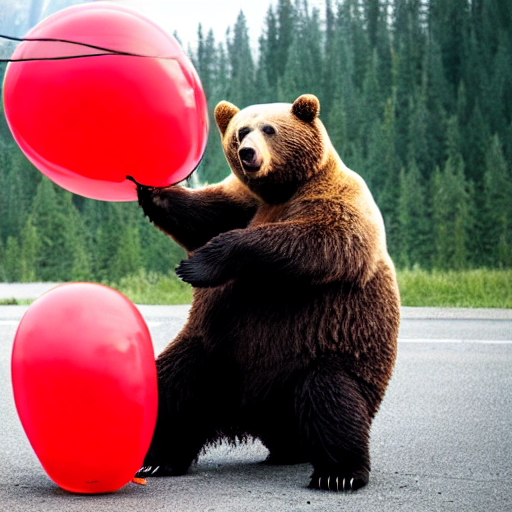} & \includegraphics[width=0.1\textwidth, height=17mm]{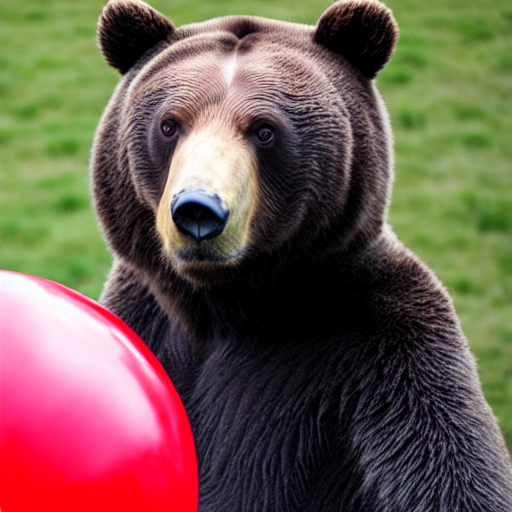} & \includegraphics[width=0.1\textwidth, height=17mm]{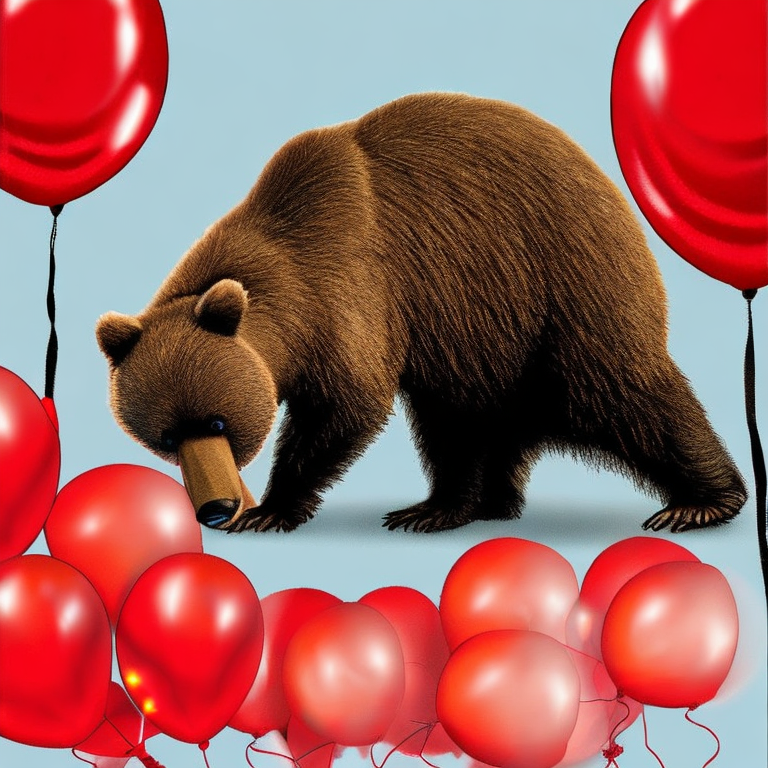} & \includegraphics[width=0.1\textwidth, height=17mm]{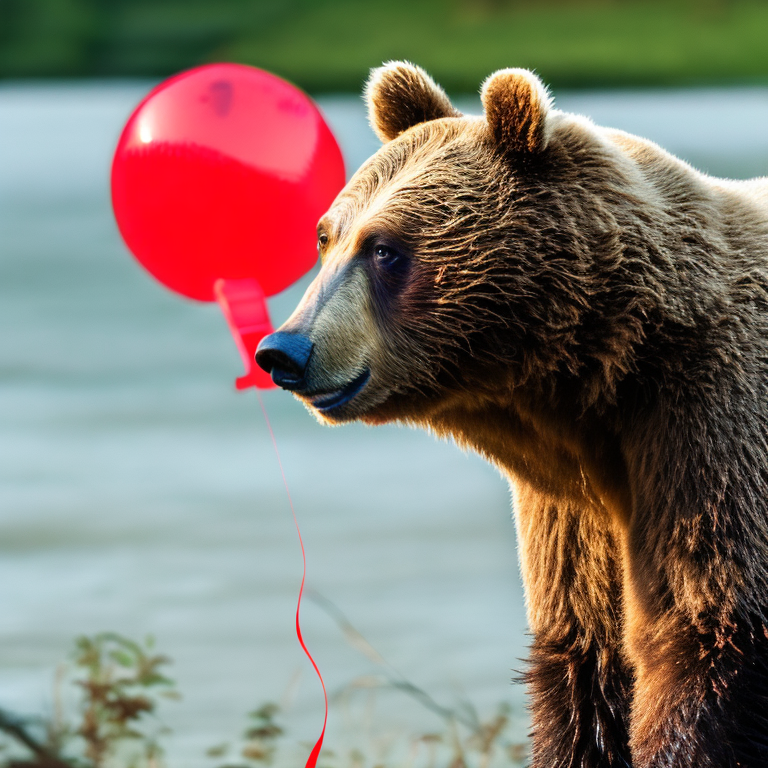} \\
250         & \includegraphics[width=0.1\textwidth, height=17mm]{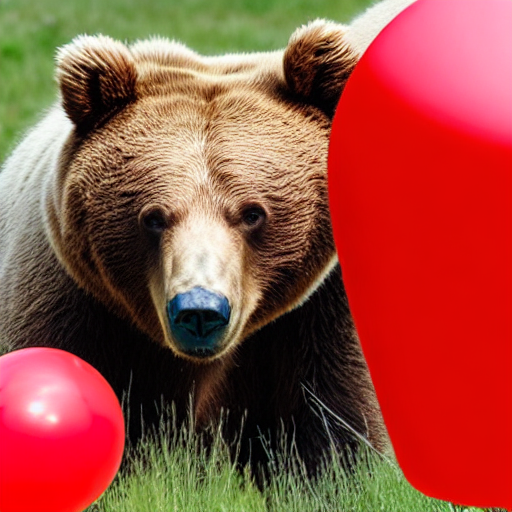} & \includegraphics[width=0.1\textwidth, height=17mm]{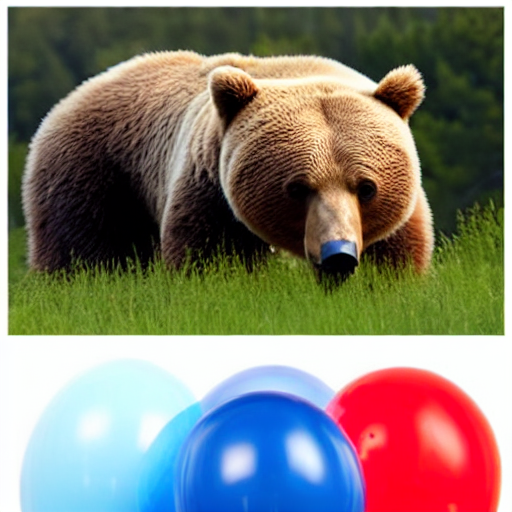} & \includegraphics[width=0.1\textwidth, height=17mm]{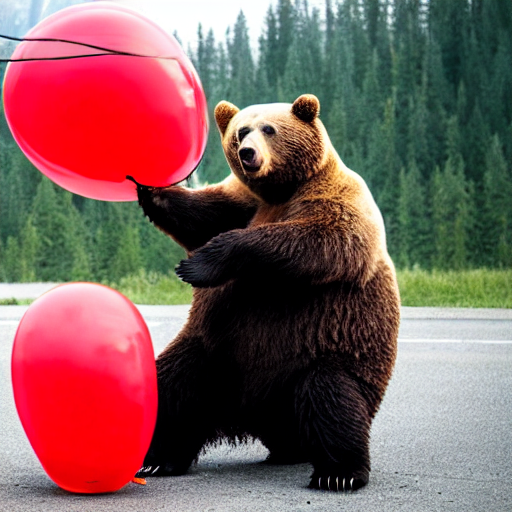} & \includegraphics[width=0.1\textwidth, height=17mm]{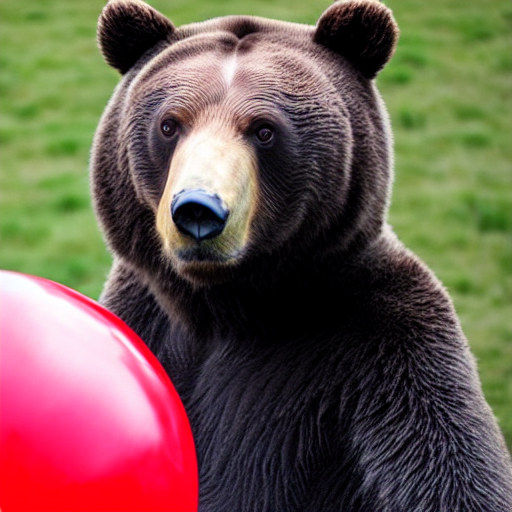} & \includegraphics[width=0.1\textwidth, height=17mm]{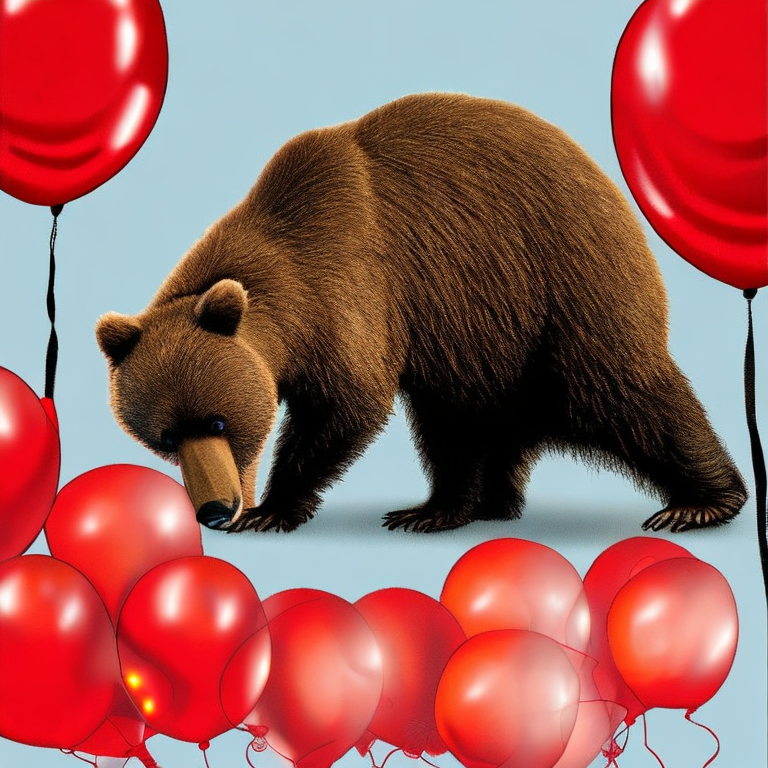} & \includegraphics[width=0.1\textwidth, height=17mm]{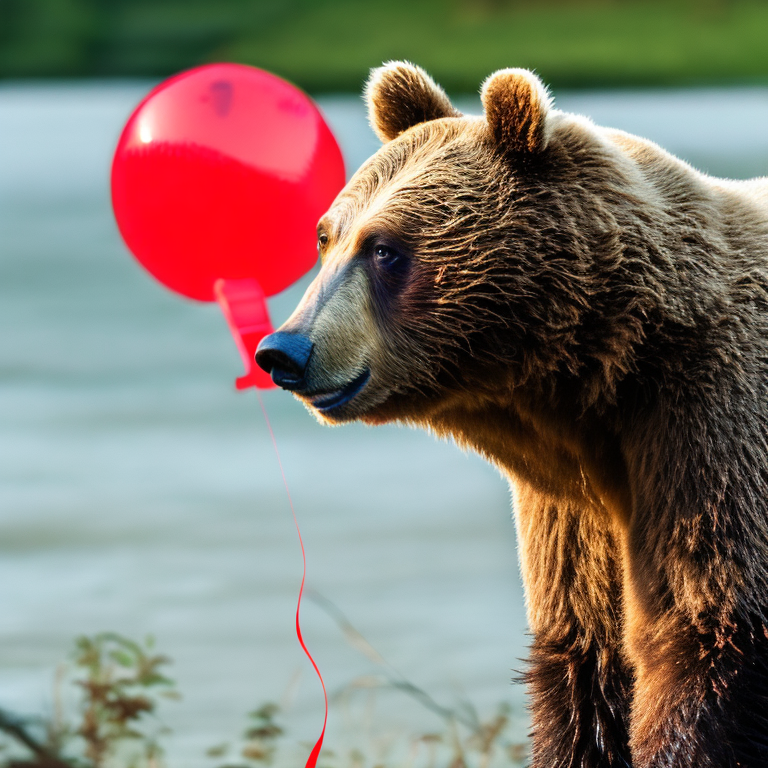} \\
300         & \includegraphics[width=0.1\textwidth, height=17mm]{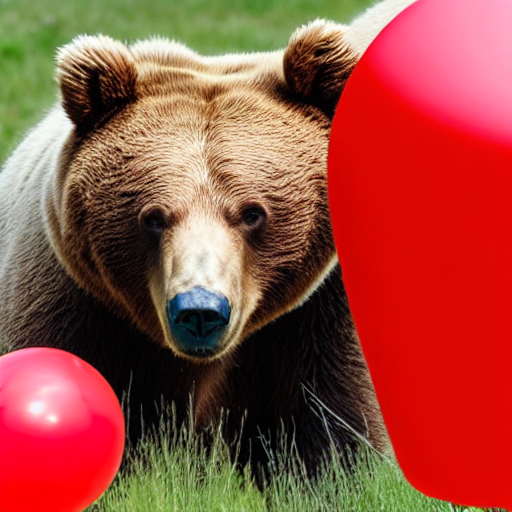} & \includegraphics[width=0.1\textwidth, height=17mm]{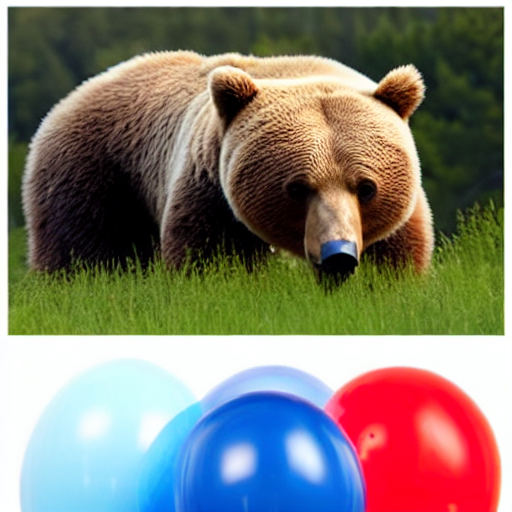} & \includegraphics[width=0.1\textwidth, height=17mm]{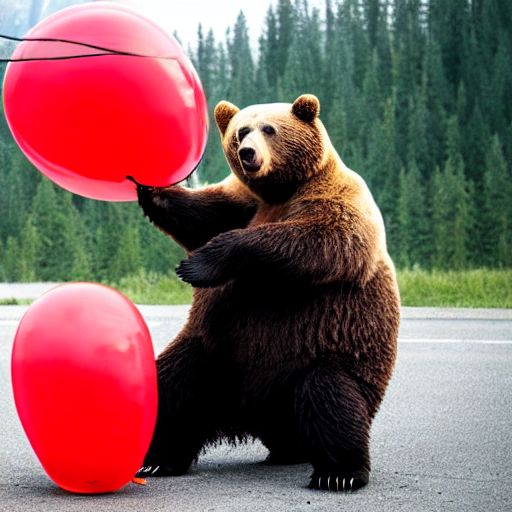} & \includegraphics[width=0.1\textwidth, height=17mm]{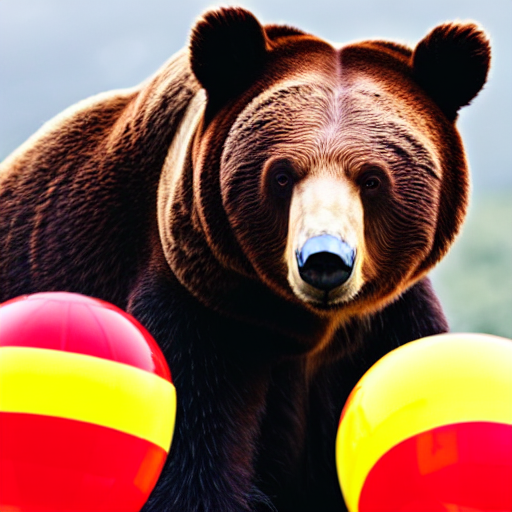} & \includegraphics[width=0.1\textwidth, height=17mm]{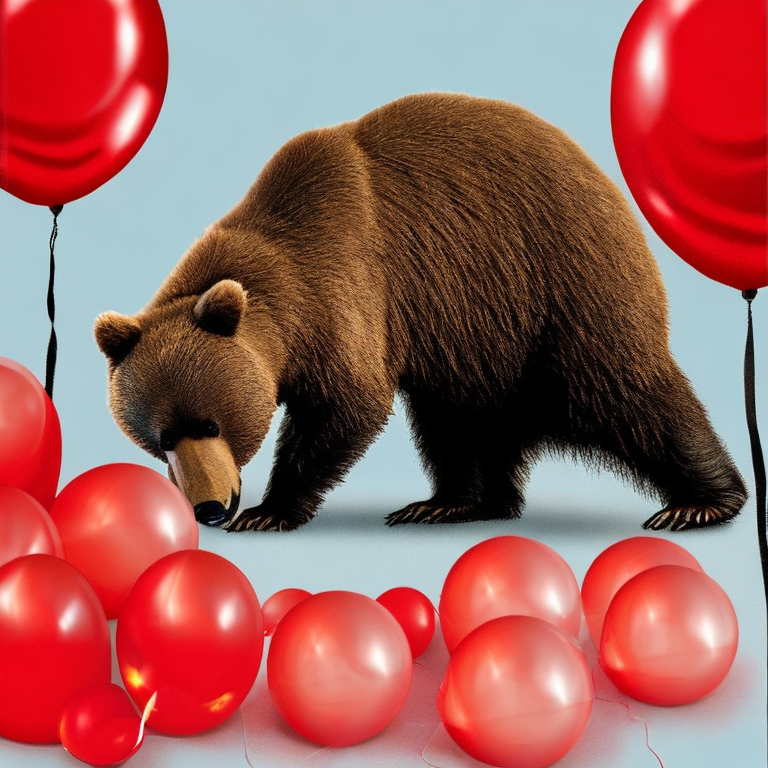} & \includegraphics[width=0.1\textwidth, height=17mm]{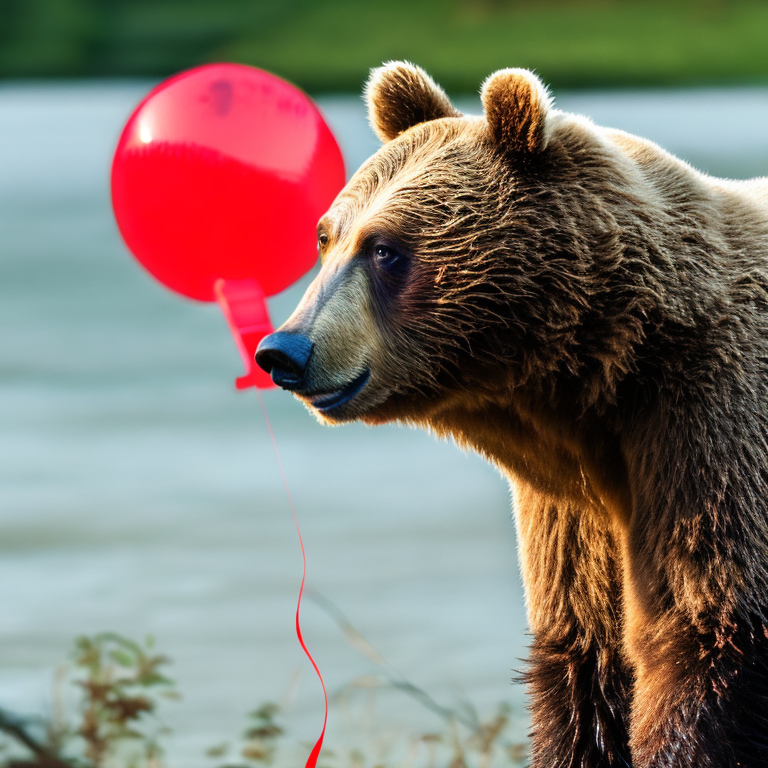} \\
\bottomrule
\end{tabular}
\end{center}
\vskip -0.1in
\end{table}


\end{document}